\documentclass[10pt,twocolumn,letterpaper]{article}

\usepackage[pagenumbers]{cvpr}

\usepackage{xspace}

\usepackage{booktabs}  \usepackage{multirow}  \usepackage{caption}   \usepackage{pifont}    \usepackage{colortbl}  \usepackage{tcolorbox} \usepackage{listings}  

\definecolor{rowhighlight}{RGB}{245,250,255}
\definecolor{niceblue}{RGB}{230,243,255}
\definecolor{reddy}{RGB}{255,238,238}
\definecolor{nicegray}{RGB}{96,96,96}

\newcommand{\cmark}{\centering\checkmark}

\definecolor{cvprblue}{rgb}{0.21,0.49,0.74}
\usepackage[pagebackref,breaklinks,colorlinks,allcolors=cvprblue]{hyperref}
\usepackage{bbold}

\newcommand{\pw}[1]{}
\newcommand{\gs}[1]{}
\newcommand{\zc}[1]{}
\newcommand{\qh}[1]{}
\newcommand{\td}[1]{}

\title{Visual Reasoning through Tool-supervised Reinforcement Learning}

\author{
Qihua Dong\textsuperscript{1,2}\thanks{Work done during an internship at Amazon.} \quad
Gozde Sahin\textsuperscript{2}\quad
Pei Wang\textsuperscript{2}\quad
Zhaowei Cai\textsuperscript{2}\quad
\\
Robik Shrestha\textsuperscript{2}\quad
Hao Yang\textsuperscript{2}\quad
Davide Modolo\textsuperscript{2}\\
\textsuperscript{1}Northeastern University\quad
\textsuperscript{2}Amazon AGI
}

\begin{document}

\maketitle

\begin{abstract}
In this paper, we investigate the problem of how to effectively master tool-use to solve complex visual reasoning tasks for Multimodal Large Language Models. To achieve that, we propose a novel {\bf Tool}-{\bf s}upervised {\bf R}einforcement {\bf L}earning (ToolsRL) framework, with direct tool supervision for more effective tool-use learning. We focus on a series of simple, native, and interpretable visual tools, including zoom-in, rotate, flip, and draw point/line, whose tool supervision is easy to collect.
A reinforcement learning curriculum is developed, where the first stage is solely optimized by a set of well motivated tool-specific rewards, and the second stage is trained with the accuracy targeted rewards while allowing calling tools.
In this way, tool calling capability is mastered before using tools to complete visual reasoning tasks, avoiding the potential optimization conflict among those heterogeneous tasks. Our experiments have shown that the tool-supervised curriculum training is efficient and ToolsRL can achieve strong tool-use capabilities for complex visual reasoning tasks.
\end{abstract}

\section{Introduction}
\label{sec:intro}

Recent advances in Multimodal Large Language Models (MLLMs) have demonstrated substantial progress in text-only reasoning (thinking-with-text)~\cite{openai2024gpt4o,bai2025qwen25vl,li2024llavaonevisioneasyvisualtask,dong2025cotreferringimprovingreferring}. However, the capabilities of these models in visual reasoning (thinking-with-images) remain comparatively less explored. Specifically, text-only reasoning proves inadequate for complex visual analysis tasks, such as interpreting rotated text or localizing small objects within cluttered scenes. A promising direction for enhancing MLLM visual reasoning involves integrating visual augmentation tools (e.g., zoom-in, rotation, drawing functions). These tools can generate intermediate visual evidence to support the reasoning process~~\cite{openai2024o3,hu2024visualsketchpad,su2025openthinkimg,zhang2025chainoffocus,wu2025vilasr,zheng2025deepeyes,lai2025minio3,chen2025rrvf}. While proprietary models (e.g., OpenAI-o3 \cite{openai2024o3}) have shown success, effective and autonomous tool-use capability—specifically determining the optimal invocation strategies (how, when, and why)—remains a significant, unsolved challenge for open-source MLLMs and the broader research community.

\begin{figure}[t]
\centering
\includegraphics[width=1\linewidth]{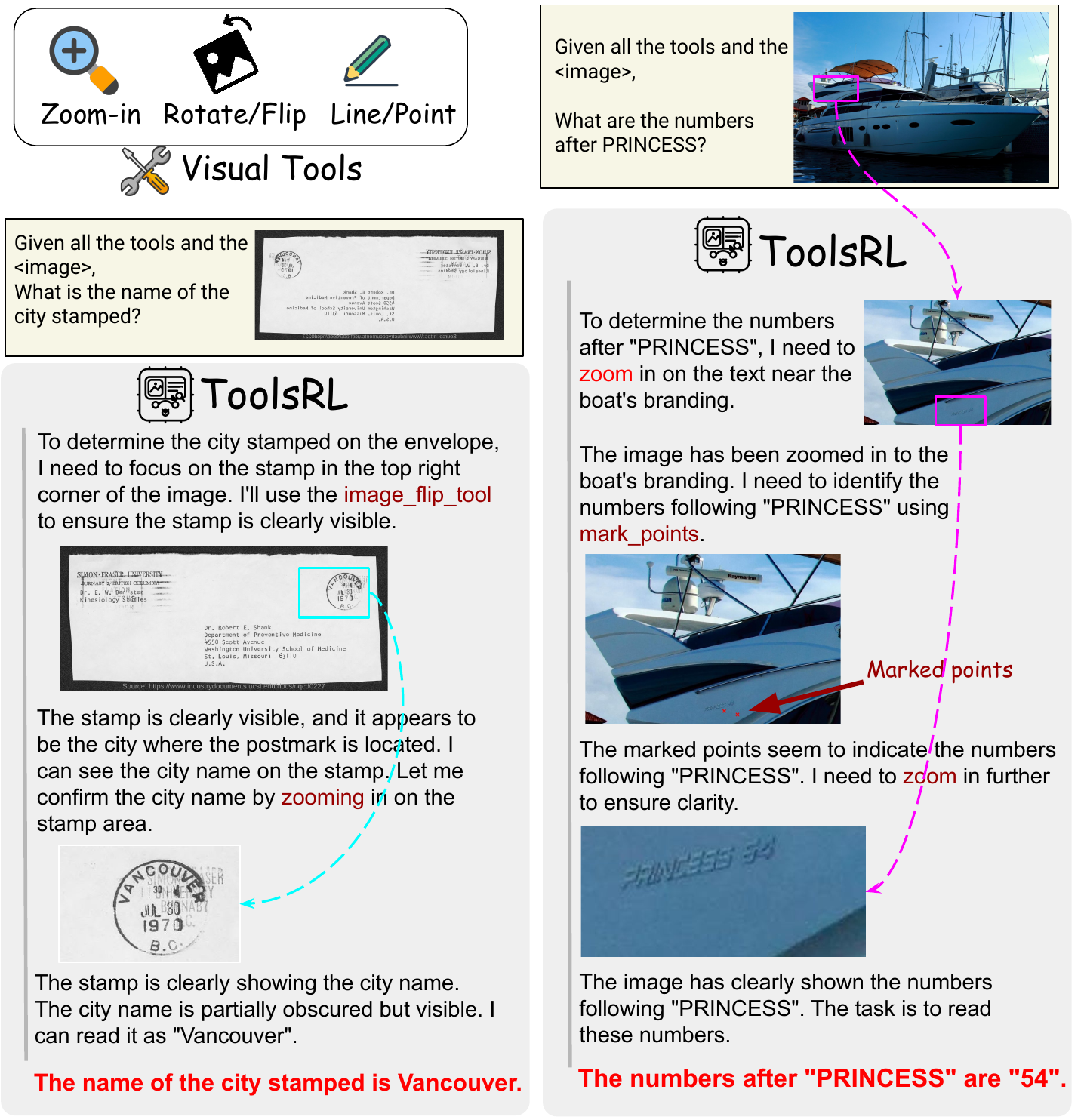}
\caption{
\textbf{Visual reasoning with ToolsRL.}
Illustrative examples of tool-supervised RL integrating various visual tools into coherent multi-step reasoning chains for different tasks.
}
\label{fig:teaser}
\end{figure}

Previous research efforts to instill tool-use capabilities primarily leveraged Supervised Fine-Tuning (SFT) on expert tool-use trajectories~\cite{su2025openthinkimg,lai2025minio3,wang2025simpleo3,zhang2025chainoffocus,wu2025vilasr}. However, this approach faces significant scalability challenges~\cite{jin2025rlpanacea,zhang2025chord} due to the substantial manual effort required to construct high-quality expert trajectories (typically generated via prompting stronger reasoning models). Furthermore, recent findings indicate that SFT trajectories require rigorous curation to prevent overfitting and maintain generalization capacity~\cite{jin2025rlpanacea,zhang2025chord}.
A more scalable alternative is Reinforcement Learning (RL) methods, such as GRPO~\cite{shao2024deepseekmath}, which enable models to explore and acquire tool-use strategies without relying on expert SFT trajectory data~\cite{zheng2025deepeyes,wu2025vtoolr1,kumar2025reinforcingvlmtools,chen2025rrvf}.
However, current RL-based methods are constrained by conventional reward design. Specifically, their reward functions often rely solely on the final task outcome~\cite{wu2025vtoolr1} or provide only generic encouragement for any tool invocation~\cite{zheng2025deepeyes,kumar2025reinforcingvlmtools,chen2025rrvf}, lacking explicit guidance on the optimal timing and execution of tool usage. Consequently, such simple rewards lead to inefficient tool-use training. Models often exhibit infrequent tool invocation (typically fewer than one per episode) and struggle to establish the coherent, multi-step tool-use chains necessary for complex visual reasoning.

To overcome the inherent limitations of both SFT-based and existing RL-based tool-use training pipelines, we introduce {\bf Tool}-{\bf s}upervised {\bf R}einforcement {\bf L}earning (ToolsRL), a novel framework that integrates standard task accuracy rewards with direct \emph{tool supervision} during the RL training process. This tool supervision provides explicit feedback on tool invocation, directly addressing the critical lack of targeted guidance observed in prior RL-based methods.
Specifically, we focus on a set of simple, native, and interpretable visual tools, including \emph{zoom-in}, \emph{rotate}, \emph{flip}, \emph{draw line}, and \emph{draw point}, whose ground-truth supervision is easy to collect. For example, the bounding box of the object of interest is utilized as the supervision signal for \emph{zoom-in}, and the underlying rotation degree of the image is for \emph{rotate}. To use these tool supervision signals during RL training, we have designed a suite of novel, well-motivated, and tool-specific reward functions, which encourage right and effective tool invocation.

In general, all rewards are supposed to be optimized jointly during RL training.
However, we observed that optimizing both tool-supervised and task accuracy rewards together in a single stage is ineffective, as models frequently defaulted to text-only reasoning. This failure to establish a critical link between tool manipulation and successful task completion motivates our two-stage training curriculum. First, the \textit{Tool Supervision Stage} focuses solely on mastering tool manipulation, with the proposed tool-supervised rewards.
Second, the \textit{Task Accuracy Stage} is optimizing the task accuracy rewards only, but allowing calling tools to generate intermediate visual evidences for complex visual reasoning tasks.
This curriculum design avoids the potential optimization conflict of the heterogeneous tool and accuracy rewards together, and enables the model to master different skills stage by stage.

In our experiments, ToolsRL has shown strong empirical performance across various tasks demanding visual reasoning capability, e.g. rotated document analysis, high-resolution image understanding, and chart comprehension, as visualized in Figure~\ref{fig:teaser}. Our contributions are:
\begin{itemize}
\item We propose Tool-supervised Reinforcement Learning (ToolsRL), a simple yet effective two-stage curriculum that enables the model to master tool-use for complex visual reasoning tasks.
\item We design well-motivated tool-supervised reward functions for a series of visual tools, which only need a small amount of easily accessible tool annotations, eliminating the need for expensive expert trajectories.
\item We provide comprehensive empirical validation across diverse tool-use tasks, demonstrating that ToolsRL yields stable training dynamics, strong accuracy, and enhanced generalization.
\end{itemize}
 \section{Related Work}
\label{sec:related}

\paragraph{RL for Multimodal Reasoning Without Explicit Tools.}

Recent works apply reinforcement learning to multimodal language models using accuracy and format rewards~\cite{huang2025visionr1,tan2025reasonrft,yao2025r1sharevl,ni2025pointrft,yang2025lookback}, often initialized with multimodal chain-of-thought (CoT) or grounded rationale. These methods improve performance on tasks such as math, document, and chart understanding. Representative approaches include Vision-R1 \citep{huang2025visionr1}, which uses CoT cold start with GRPO; Reason-RFT and R1-ShareVL \citep{tan2025reasonrft,yao2025r1sharevl}, which stabilize and diversify trajectories under RL; Point-RFT \citep{ni2025pointrft}, which learns visually grounded rationales before RL; and Look-Back \citep{yang2025lookback}, which re-focuses during reasoning without callable tools. While effective for single-round text outputs, these methods cannot leverage intermediate visual manipulations for multi-step visual reasoning.

\paragraph{Visual Tool-Use in Multimodal Language Models}

To overcome the limitation that arises from lacking explicit tool guidance, a complementary line of work exposes callable visual tools (e.g., zoom-in, draw) to multimodal models~\citep{hu2024visualsketchpad,su2025openthinkimg,zhang2025chainoffocus,wu2025vilasr,zheng2025deepeyes,lai2025minio3,chen2025rrvf}, with training recipes ranging from training-free prompting ~\citep{hu2024visualsketchpad} to SFT-only~\citep{wang2025simpleo3}, SFT-then-RL~\cite{su2025openthinkimg,zhang2025chainoffocus,wu2025vilasr}, and RL-only~\cite{zheng2025deepeyes,chen2025rrvf} approaches.
In \textbf{Training-free}, for example, Visual Sketchpad treats drawing as an action interface, improving localization and counting without finetuning, but performance is limited and relies on strong base models~\citep{hu2024visualsketchpad}.
\textbf{SFT-only} models are finetuned with supervised trajectories of expert tool usage, learning explicit tool invocation patterns but without reinforcement feedback. Simple o3 is a typical example, which interleaves executable operations with a curated tools dataset~\citep{wang2025simpleo3}. However, this type of methods relies on manually curated expert trajectories that are expensive to obtain and task-specific, limiting scalability.
In \textbf{SFT-then-RL}, models first learn tool-use behavior through supervised fine-tuning and are then refined with reinforcement learning to balance tool effectiveness and final task accuracy. It is a dominant tool-use thread and there are many works on it. \emph{OpenThinkIMG} standardizes tool APIs and mixes answer-quality rewards with intermediate tool-output signals \citep{su2025openthinkimg}; \emph{Chain-of-Focus} learns adaptive zoom strategies \citep{zhang2025chainoffocus}; \emph{Mini-o3} scales to longer multi-turn visual search \citep{lai2025minio3}; and \emph{ViLaSR} uses drawing primitives in a three-stage pipeline with an RL phase \citep{wu2025vilasr}.
The same as SFT-only approaches, they all require substantial data effort to construct supervised demonstrations before reinforcement learning can begin. \textbf{RL-only} models learn tool-use strategies entirely through reinforcement learning from reward signals, without any supervised demonstrations, promoting scalability and generation. For instance, DeepEyes~\cite{zheng2025deepeyes} learns tool policies end-to-end, and RRVF~\cite{chen2025rrvf} uses render–execute–judge feedback.
Simultaneously acquiring fine-grained tool control and optimizing task objectives from sparse rewards remains highly challenging.
Our work directly addresses this challenge by incorporating tool supervision during RL.
 \begin{figure}[t]
\centering
\includegraphics[width=\linewidth]{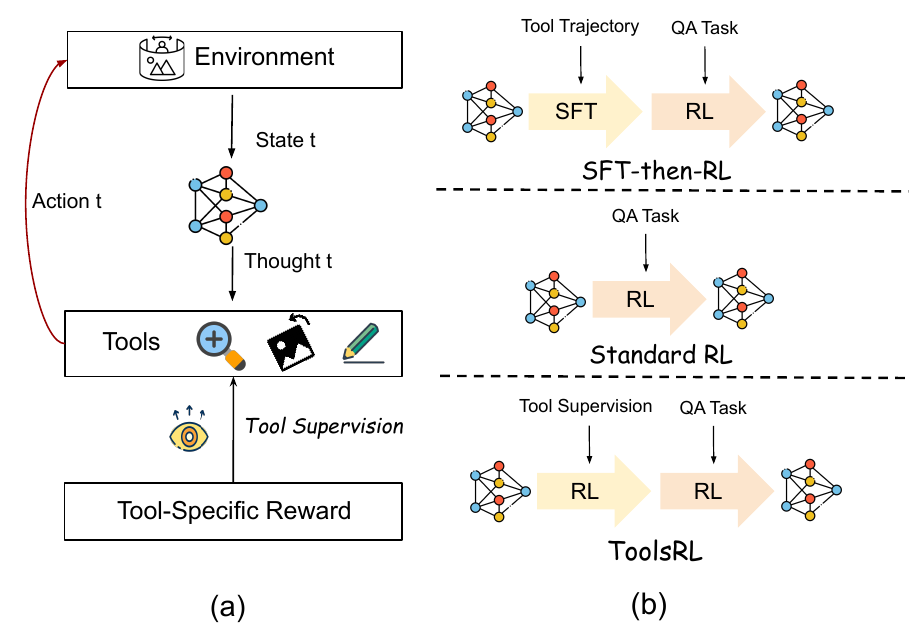}
\caption{
\textbf{Overview of Tool-supervised Reinforcement Learning (ToolsRL).}
(a) \emph{ToolsRL} includes tool-specific rewards that supervises tool usage.
(b) Unlike SFT-then-RL and standard RL, \emph{ToolsRL} injects tool supervision before training on QA tasks.
}
\label{fig:method}
\end{figure}

\section{Method}
\label{sec:method}

Figure~\ref{fig:method} provides an overview of the proposed ToolsRL framework and its two-stage training curriculum.

\subsection{Problem Formulation}

We cast visual tool use as a finite-horizon sequential decision process. At each turn $t$, the agent observes the state $s_t$, consisting of the input question, the current image, and the trajectory so far, and selects an action $a_t$. The action space includes (1) calling a visual tool with a specific argument, which produces a new image and advances to turn $t{+}1$, or (2) outputting a final answer to terminate the episode.
At each turn, the agent may apply tools to any image in the trajectory history. Each turn permits at most one tool call and therefore yields at most one new image.

The goal is to learn a policy $\pi_\theta(a_t \mid s_t)$ that maximizes expected return:
\begin{equation}
\resizebox{0.9\linewidth}{!}{$\displaystyle
\max_{\theta}\; J(\theta)\;=\; \mathbb{E}_{\tau \sim \pi_\theta}\!\left[\sum_{t=1}^{T} r(s_t, a_t)\right], \quad a_T=\texttt{<answer>}
$}
\end{equation}
where $r(s_t, a_t)$ is the per-step reward at turn $t$, $\tau=(s_1,a_1,\dots,s_T,a_T)$ is a trajectory, and $T$ is stopping time.

\subsection{Tool Supervision for RL}
\subsubsection{Tool Suite and Tasks}
\label{sec:tool-supervision-settings}
We use three core tools that cover the essential visual operations our framework requires: \textbf{Zoom-in} (crop to a bounding box and resize), \textbf{Rotate/Flip} (90°/180°/270° rotations or horizontal/vertical flips), and \textbf{Draw} (overlay horizontal/vertical lines or points on the image). Together, they span region selection, orientation correction, and coordinate-based annotation. Note that we focus on native tool-calling in this work, and do not consider calling external tools (e.g. standalone segmentation models).

Unlike tool-use SFT, which imitates whole trajectories including all textual reasoning and tool uses, tool supervision is more flexible and scalable. We supervise each task type with ground-truth specific to that task.
Given a base visual question-answer dataset, we prepare ground-truth supervision for each of our tasks as follows:

\begin{itemize}
  \item \textbf{Zoom-in tasks:} We use the ground-truth bounding boxes of objects or regions from the question to define the target crop area, and record the corresponding zoom-in operation as the ground-truth tool use.
  These bounding boxes are obtained from datasets that provide object-level annotations, and serve as supervision for the tool’s spatial localization behavior.

  \item \textbf{Rotate/Flip tasks:} We augment images with random rotations and flips, recording the inverse transformation as ground-truth. 

  \item \textbf{Draw tasks:} We synthesize chart-style questions and define ground-truth tool use as drawing a horizontal/vertical line to read a point's $x$/$y$ value, or placing points to support counting or marking (read-value and compare-and-count tasks).
\end{itemize}

\subsubsection{Tool-supervised Reward Design}
\label{sec:tool-supervised-rewards}
Building on the ground-truth tool supervision we highlight for each task above, we design rewards to evaluate the tools invoked by the model at each step. We adopt a per-state view: at state $s_t$ with current image $I_t$, the per-state reward $R_{\text{task}}(s_t, \mathcal{G}^{\text{task}})$ is computed from all tools applied at $s_t$ using the ground-truth set $\mathcal{G}^{\text{task}}$ for that sample.

\paragraph{Zoom-in: Modified F1 reward.}
For zoom-in, the model must localize visual elements by predicting bounding boxes in tool calls, which lets it crop and resize to target regions in the image. We define a pixel-level modified F1-style overlap metric, ModF1, to evaluate zoom-in tool calls. True positives (TP), false positives (FP), and false negatives (FN) are computed from the intersection-over-union (IoU) between the predicted box mask $b$ and the ground-truth box mask $g$ at the pixel level:

\begin{equation}
    \mathrm{ModF1}\bigl(b, g\bigr) \,=\, \frac{2\,\mathrm{TP}}{2\,\mathrm{TP} + w_{\text{fp}}\,\mathrm{FP} + w_{\text{fn}}\,\mathrm{FN}}\,,
\end{equation}
where $g$ is a ground-truth bounding box and $b$ is the zoom-in box in the tool call. Factors $w_{\text{fp}}$ and $w_{\text{fn}}$ are the weighting coefficients we apply for FP and FN. Because zoom-in is not a strict grounding task, spurious zooms (FP) are far less harmful than missing the target region (FN). This is why we introduce the coefficients $w_{\text{fp}}$ and $w_{\text{fn}}$, emphasizing recall over precision in our reward design. In our final framework, we use $w_{\text{fp}}{=}0.1$ and $w_{\text{fn}}{=}1.0$ to reflect this asymmetry.

We compute the per-state reward by matching the predicted zoom-in box $b$ at state $s_t$ to the best ground-truth box in $\mathcal{G}^{\text{zoom-in}}$:
\begin{equation}
R_{\text{zoom-in}}(s_t, \mathcal{G}^{\text{zoom-in}}) \,=\, \max_{g_i \in \mathcal{G}^{\text{zoom-in}}}\; \mathrm{ModF1}\bigl(b, g_i\bigr).
\end{equation}

\paragraph{Rotate/Flip: Orientation reward.}
For rotate/flip tasks, $\mathcal{G}^{\text{rotflip}}$ is the canonical orientation for the original input image; equivalently, it defines the target orientation $o^*$ for $I_t$.
Since we evaluate \emph{only} calls on the current image $I_t$, we use a binary per-state reward:
\begin{equation}
R_{\text{rotflip}}(s_t, \mathcal{G}^{\text{rotflip}}) \,=\, \mathbb{1}\!\left[\text{o}\!\left(I_t\right) = o^*\right] \in \{0,1\},
\end{equation}
where $o(I_t)$ is the orientation for the current image, and $o^*$ is the target orientation.

\paragraph{Draw: Unified coordinate-based reward.}

For tasks requiring precise spatial reasoning, the model draws lines or points at specific coordinates.
We use a single margin-based score for both primitives (line and point).
Given a predicted primitive $p$ and a ground-truth primitive $p^*$, we compute a similarity score as:
\begin{equation}
 s(p, p^*) \,=\, \max\!\Bigl(0,\, 1 - \tfrac{d(p,p^*)}{T_{p^*}}\Bigr),
\end{equation}
where $d(\cdot)$ is the primitive-appropriate distance function and $T_{p^*} \in \{T_x,T_y,T_p\}$ is the tolerance for the matched primitive type (x-axis line, y-axis line, or point).

For lines along axis $a{\in}\{x,y\}$, Let $c$ denote the predicted line coordinate along axis $a$, and $c_a^*$ the ground truth coordinate. Then,
\begin{equation}
  d_{\text{line}} = |c - c_a^*|, \quad T_x = W/4, \quad T_y = H/4,
\end{equation}
 where $W$ and $H$ are the image width and height, respectively.
For points, Let $p = (x_p, y_p)$ and $p^* = (x_p^*, y_p^*)$ denote the predicted and ground-truth points. Then,

\begin{flalign}
    d_{\text{point}} &= \sqrt{(x_p - x_p^*)^2 + (y_p - y_p^*)^2}, \\
    T_p &= \sqrt{(W/4)^2 + (H/4)^2}.
\end{flalign}

Intuitively, $s(p, p^*)$ is 1 when the predicted primitive exactly matches the ground truth, and decreases linearly to 0 as the prediction reaches the tolerance threshold.

For the per-state reward, let $\mathcal{C}_t^{\text{draw}}$ be the set of predicted primitives (all required lines and/or points) produced at $s_t$, and let $\mathcal{G}^{\text{draw}}$ be the corresponding ground-truth primitives that contain line coordinates and/or point locations. We compute a similarity score between predictions and ground truth using Hungarian matching. $S_{\text{TP}}$ is defined as the sum of per-primitive similarities $s(p,p^*)$ for the optimal one-to-one matching between $\mathcal{C}_t^{\text{draw}}$ and $\mathcal{G}^{\text{draw}}$. So $S_{\text{TP}}$ maximizes the total similarity.

Finally, we define the final F1-style reward for draw tasks jointly as:
\begin{equation}
 R_{\text{draw}}(s_t, \mathcal{G}^{\text{draw}}) \,=\, \frac{2\,S_{\text{TP}}}{\,|\mathcal{C}_t^{\text{draw}}| + |\mathcal{G}^{\text{draw}}|\,}\,,
\end{equation}
which seamlessly unifies lines and points without separate reward formulas.

\subsection{The Two-Stage Tool-supervised Curriculum}
In our framework, we adopt a two-stage curriculum that decouples the mechanics of tool use from answer prediction. In \textbf{Stage~1 (Tool Supervision)}, the model learns to operate the toolbox accurately and consistently using ground-truth derived rewards; in \textbf{Stage~2 (Task Accuracy)}, it learns to produce the correct final answer while freely leveraging the tools it has mastered.

\subsubsection{Stage 1: Tool-supervision}
\label{sec:tool-supervision}

This stage optimizes task-specific tool-accuracy rewards computed directly from the tool calls.
The model is prompted to explicitly identify and manipulate visual elements using the available tools.
For example, a zoom-in tool-supervised question might ask:
\begin{quote}\small
    \textbf{Tool-supervised question.} The sentence is: ``What does the label on the bottom right corner of the yellow fabric on the fourth shelf of the cabinet on the left say''. First, identify the visual elements (objects or text) referenced in the sentence. Then, use zoom-in to locate each element in the image and report which zoom call finds it (index starts from 1). Example: if ``bike'' is found on the 3rd image and ``person'' on the 8th, answer: \texttt{<answer>{"bike": 3, "person": 8}</answer>}
\end{quote}

Sample tool-supervised questions for other tasks are provided in the Suppl.

For our final Stage 1 reward, we define two reward components to balance exploration and task-awareness: a \emph{global} tool reward $R^{\text{global}}_{\text{tool}}$ that evaluates the complete tool trace (encouraging broad exploration), and an \emph{answer-conditioned} tool reward $R^{\text{answer}}_{\text{tool}}$ that evaluates only the tool calls applied to the image referenced in the model's \verb|<answer>| (enabling awareness of effective tool use).

Global-only rewards encourage exploration but may reward irrelevant steps; while answer-only rewards improve relevance but hinder discovery. Using both ($R^{\text{global}}_{\text{tool}}$ and $R^{\text{answer}}_{\text{tool}}$) balances exploration with task relevance and reduces inefficient tool uses.

Let $R_{\text{task}}(s_t, \mathcal{G}^{\text{task}})$ denote the appropriate per-state reward (e.g., $R_{\text{zoom-in}}$, $R_{\text{rotflip}}$, or $R_{\text{draw}}$) defined above for the tool-specific task,
and let $t_{\text{answer}}$ denote the state index referenced in the model's \verb|<answer>| tag.
We define
\begin{align}
R^{\text{global}}_{\text{tool}} &\,=\, \max_{t \in \{1,\dots,T\}} \; R_{\text{task}}(s_t, \mathcal{G}^{\text{task}}),\\
R^{\text{answer}}_{\text{tool}} &\,=\, R_{\text{task}}(s_{t_{\text{answer}}}, \mathcal{G}^{\text{task}}).
\end{align}

The final reward for Stage 1 is then defined as:
\begin{equation}
R_{\text{final, stage-1}} \,=\, \tfrac{1}{2}\bigl(R^{\text{global}}_{\text{tool}} + R^{\text{answer}}_{\text{tool}}\bigr) + R_{\text{format}},
\end{equation}
where $R_{\text{format}}$ is the format reward as defined in \citep{zheng2025deepeyes}.

\subsubsection{Stage 2: Task Accuracy}
\label{sec:task-accuracy}

In this stage, the model receives a standard QA prompt, with no tool-specific supervision applied. The model may still call tools (and typically does, at increasing rates as training progresses), but the sole objective is answer accuracy, which we measure using an LLM judge for all datasets except our synthetic chart sets. For read-value and compare-and-count tasks in our synthetic chart data, we use the normalized numerical score $s_{norm}$ (calculating the difference of our answer and ground-truth and normalizing it by the $x$/$y$ range of the chart or the number of total points).

\begin{equation}
\resizebox{.8\hsize}{!}{$R_{\text{answer}} = \begin{cases}
s_{norm}\!\left(\text{ans}, \text{ans}^*\right), & \text{task}\in\text{synth. chart}\\
\mathbb{1}_{\text{LLM-Judge}}\!\left[\text{ans} = \text{ans}^*\right], & \text{else},
\end{cases}$}
\label{eq:R_answer}
\end{equation}
where ${ans}$ is the model's final answer, ${ans}^*$ is the target answer, and $\mathbb{1}_{\text{LLM-Judge}}$ denotes a binary judgment by the LLM judge, following \cite{zheng2025deepeyes}. The final reward for this stage combines answer correctness with format compliance:

\begin{equation}
R_{\text{final, stage-2}} = R_{\text{answer}} + R_{\text{format}},
\end{equation}
where $R_{\text{format}}$ rewards adherence to the expected output structure, again following its definition in \cite{zheng2025deepeyes}.

\subsubsection{Why does curriculum learning matter?}
\label{sec:discussion}
The two-stage curriculum is crucial for effective tool learning. By decoupling tool mastery from answer accuracy, tool-supervision stage (Stage 1) allows the model to focus exclusively on learning \emph{how} to use tools correctly without the confounding pressure of producing correct answers. Once the model has internalized tool usage patterns in Stage~1, it can naturally leverage these learned capabilities in task accuracy stage (Stage~2) to improve answer accuracy. In contrast, training directly on answer accuracy from the start causes the model to prefer text-based reasoning over tool use. Examples of this can be seen in Figure~\ref{fig:comparison-case-study}.
We provide detailed ablation studies comparing our curriculum approach against combined reward training in Table~\ref{tab:ablation_curriculum}.

 \begin{table*}[htbp]
\centering
\caption{Comparison with SOTA on document understanding, spatial reasoning, and chart understanding groups. ``--'' denotes that the method was not evaluated due to the lack of open-sourced models. ``$^{*}$'' indicates that the results are evaluated by us using open-sourced weights. All methods use Qwen2.5-VL-7B as the base model.
}
\label{tab:main}
\begingroup
\renewcommand{\arraystretch}{1.25}
\resizebox{\textwidth}{!}{\begin{tabular}{l c c | *{16}{c}}
\toprule
\multirow{2}{*}{Method} & \multicolumn{2}{c|}{Tool-Use Training} &
\multicolumn{2}{c}{Document Understanding} &
\multicolumn{11}{c}{Spatial Reasoning} &
\multicolumn{3}{c}{Chart/Table Understanding} \\
\cmidrule(lr){2-3} \cmidrule(lr){4-5} \cmidrule(lr){6-16} \cmidrule(lr){17-19}
& \multirow{2}{*}{SFT} & \multirow{2}{*}{RL} &
\multirow{2}{*}{DocVQA-RF} & \multirow{2}{*}{InfoVQA-RF} &
\multirow{2}{*}{InfoVQA-Res} & \multicolumn{3}{c}{V-Star} & \multicolumn{3}{c}{HR-Bench 4K} & \multicolumn{3}{c}{HR-Bench 8K} & \multirow{2}{*}{VisualProbe} &
\multirow{2}{*}{CharXiv} & \multirow{2}{*}{ChartQA-Pro} & \multirow{2}{*}{TableVQA} \\
& & &
 &  &
 & V$^{*}$-S & V$^{*}$-C & Avg & HR-4K-S & HR-4K-C & Avg & HR-8K-S & HR-8K-C & Avg & &
 &  &  \\
\midrule
Qwen2.5-VL~\cite{bai2025qwen25vl}  & - & - & 50.2$^{*}$ & 53.8$^{*}$ & 50.9$^{*}$ & 78.2$^{*}$ & 73.6$^{*}$ & 75.9$^{*}$ & 83.8$^{*}$ & 56.9$^{*}$ & 70.4$^{*}$ & 78.8$^{*}$ & 51.8$^{*}$ & 65.3$^{*}$ & 28.4$^{*}$ & 41.2$^{*}$ & 31.7$^{*}$ & 66.2$^{*}$ \\
Point-RFT~\cite{ni2025pointrft}  & \checkmark & \checkmark & -- & -- & -- & -- & -- & -- & -- & -- & -- & -- & 36.20 & -- & -- \\
ZoomEye~\cite{shen2025zoomeye} & - & - & -- & -- & -- & 93.9 & 85.5 & 89.7 & 84.3 & 55.0 & 69.7 & 88.5 & 50.0 & 69.3 & -- & -- & -- & -- \\
Simple o3~\cite{wang2025simpleo3} & \checkmark & - & -- & -- & -- & -- & -- & 90.4 & -- & -- & 76.2 & -- & -- & -- & -- & 41.8 & -- & -- \\
Pixel-Reasoner~\cite{su2025pixelreasoner} & \checkmark & \checkmark & -- & -- & -- & -- & -- & 86.3 & -- & -- & 74.0 & -- & -- & 66.9 & 38.9 & -- & -- & -- \\
Mini-o3~\cite{lai2025minio3} & \checkmark & \checkmark & 52.9$^{*}$ & 31.3$^{*}$ & 58.2$^{*}$ & -- & -- & 88.2 & -- & -- & \textbf{77.5} & -- & -- & 73.3 & \textbf{55.1} & 37.3$^{*}$ & 32.9$^{*}$ & 56.5$^{*}$ \\
DeepEyes~\cite{zheng2025deepeyes} & - & \checkmark & 61.3$^{*}$ & 59.7$^{*}$ & 59.5$^{*}$ & 91.3 & 88.2 & 89.8 & \textbf{91.3} & 59.0 & 75.2 & 86.8 & 58.5 & 72.7 & 41.6 & 38.5$^{*}$ & 37.2$^{*}$ & 67.4$^{*}$ \\
\midrule
\rowcolor{reddy}ToolsRL (ours) & - & \checkmark & \textbf{77.3} & \textbf{61.4} &  \textbf{71.0} & \textbf{95.6} & \textbf{89.4} & \textbf{92.5} & 91.2 & \textbf{60.6} & 75.9 & 88.1 & 58.3 & 73.2 & {46.5} & \textbf{43.5} & \textbf{38.8} & \textbf{70.2} \\
\bottomrule
\end{tabular}}
\endgroup
\vspace{0.5em}
\end{table*}
 
\begin{table*}[htbp]
\centering
\caption{Ablation results for the components of our framework that cover different reward strategies as well as training strategies (with or without curriculum).}
\label{tab:ablation_curriculum}
\resizebox{\textwidth}{!}{\newcommand{\rotrow}[1]{\rotatebox[origin=c]{90}{\makebox[1.2cm][c]{\footnotesize #1}}}
\begin{tabular}{l c c c c c | cc | cccccc | ccc}
\toprule

\multirow{2}{*}{Method} &
\multirow{2}{*}{$R_{\text{answer}}$} &
\multirow{2}{*}{$R_{\text{tool\_cond}}$} &
\multirow{2}{*}{$R^{\text{global}}_{\text{tool}}$} &
\multirow{2}{*}{$R^{\text{answer}}_{\text{tool}}$} &
\multirow{2}{*}{Curric.} &
\multicolumn{2}{c}{Document Understanding} &
\multicolumn{6}{c}{Spatial Reasoning} &
\multicolumn{3}{c}{Chart/Table Understanding} \\
\cmidrule(lr){7-8} \cmidrule(lr){9-14} \cmidrule(lr){15-17}
& & & & & &
DocVQA-RF & InfoVQA-RF &
InfoVQA-Res & VStar-S & VStar-C & HR-4K-S & HR-4K-C & VisualProbe &
CharXiv & ChartQA-Pro & TableVQA \\
\midrule
Qwen2.5-VL-7B & - & - & - & - & - & 50.2$^{*}$ & 53.8$^{*}$ & 50.9$^{*}$ & 78.2$^{*}$ & 73.6$^{*}$ & 83.8$^{*}$ & 56.9$^{*}$ & 28.4$^{*}$ & 41.2$^{*}$ & 31.7$^{*}$ & 66.2$^{*}$  \\
\midrule
Ans. Reward  & \cmark &  &  &  &  & 62.6 & 60.0 & 60.2 & 93.0 & 89.5 & 92.2 & 57.9 & 41.9 & 42.0 & 37.6 & \textbf{70.6} \\
Tool-Cond. \& Ans. Reward & \cmark & \cmark &  &  &  & 71.1 & 59.1 & 62.5 & 92.2 & 80.3 & 84.3 & 57.4 & 44.1 & 43.0 & 35.2 & 68.3 \\
Tool-Sup \& Ans. Reward & \cmark &  & \cmark & \cmark &  & 58.1 & 60.9 & 55.7
& 90.2 & 89.5 & 87.8 & 53.4 & 41.4 & 41.6 & 35.9 & 70.5 \\
\midrule
Global Tool-Sup only & \cmark &  & \cmark &  & \cmark & 60.3 & 58.4 & 58.4 & 95.1 & 88.4 & 89.8 & 56.0 & 43.4 & \textbf{44.1} & \textbf{40.4} & 69.9 \\
Answer Tool-Sup only & \cmark &  &  & \cmark & \cmark & 65.4 & 58.7 & 61.4 & 92.3 & \textbf{90.1} & 90.1 & 57.7 & 39.7& 43.3 & 37.7 & 68.8 \\
\rowcolor{reddy} ToolsRL (ours) & \cmark &  & \cmark & \cmark & \cmark & \textbf{77.3} & \textbf{61.4} &  \textbf{71.0} & \textbf{95.6} & 89.4 & \textbf{91.2} & \textbf{60.6} & \textbf{46.5} & 43.5 & 38.8 & 70.2 \\
\bottomrule
\end{tabular}}
\end{table*}
 \begin{table}[t]
\centering
\caption{Ablation of key design choices in tool reward formulation across task types. Columns list the key components that differ. ``Acc.'' denotes accuracy and ``T.'' denotes the average number of tool calls.}
\label{tab:ablation_components}
\resizebox{\columnwidth}{!}{\begin{tabular}{c cc| c cc| c cc}
\toprule
 Zoom-in & \multicolumn{2}{c|}{VisualProbe} & Rot./Flip & \multicolumn{2}{c|}{DocVQA-RF} & Draw & \multicolumn{2}{c}{ChartQA-Pro} \\
 \cmidrule(lr){2-3} \cmidrule(lr){5-6} \cmidrule(lr){8-9}
$w_{\mathrm{fp}}$ & Acc. & T. & data mix & Acc. & T. & reward type & Acc. & T. \\
\midrule
$w_{\mathrm{fp}}{=}1$ & 42.9 & 2.13 & aug + orig. & 67.1 & 6.98 & discrete reward & 37.9 & 2.43 \\
$w_{\mathrm{fp}}{=}0.1$ & 46.3 & 3.20 & aug only & 79.4 & 4.26 & cont. reward & 39.1 & 2.65 \\
\bottomrule
\end{tabular}}
\end{table}

\section{Experiments}
\label{sec:experiments}

\subsection{Settings}

\paragraph{Training datasets.}
To train and evaluate tool-use capabilities of ToolsRL under diverse visual reasoning scenarios, we curate a corpus covering document understanding, spatial reasoning, and chart understanding. Examples are provided in the supplementary material. Specifically,
\begin{itemize}[leftmargin=*, topsep=2pt, itemsep=1pt]
    \item \textbf{Document understanding:} 3k samples are randomly selected from DocVQA~\cite{mathew2021docvqa} and augmented with rotation and flip transformations as a training set.
    \item \textbf{Spatial reasoning:} 6k samples from SealVQA~\cite{wu2024vstar} and 8k high-resolution samples from Visual Probe~\cite{lai2025minio3} are used to train the model on fine-grained localization and spatial understanding.
    \item \textbf{Chart/table understanding:} 2k samples from ChartQA~\cite{masry2022chartqa} and 2k samples from ArxivQA~\cite{li-etal-2024-multimodal-arxiv} are utilized, complemented by our synthetic datasets Read-Value (2k samples) and Compare-and-Count (4k samples) where the model reads the x/y values of a point or count the number of points that satisfy a condition.
\end{itemize}

\noindent All datasets mentioned above are used during both Stage 1 and Stage 2 training, with the exception of ChartQA and ArxivQA, which are omitted from Stage 1 due to lack of ground-truth annotations for effective tool-supervision.

\paragraph{Evaluation datasets.}
We evaluate our method on benchmarks spanning the same three domains as the training set:
\begin{itemize}[leftmargin=*, topsep=2pt, itemsep=1pt]
    \item \textbf{Document understanding:} We use DocVQA~\cite{mathew2021docvqa} and InfoVQA~\cite{mathew2022infographicvqa} and augment their test sets with rotation and flip transformations to form DocVQA-RF and InfoVQA-RF. Rotations ($90^\circ$, $180^\circ$, or $270^\circ$) and flips (vertical or horizontal) are sampled uniformly and applied to the image with 0.7 probability.
    \item \textbf{Spatial reasoning:} HR-Bench~\cite{wang2025hrbench} and V-Star~\cite{wu2024vstar} are used for single- and cross-image high-resolution perception evaluation. Visual Probe~\cite{lai2025minio3} is used with easy, medium, and hard splits. We also construct InfoVQA-Res~\cite{mathew2022infographicvqa}, by selecting high-resolution images (max edge length $>$ 1024 pixels) and resizing them to  $ \leq 512\times512$ pixels, to evaluate the model’s reasoning performance on high-resolution infographics.
    \item \textbf{Chart/table understanding:} Evaluation is conducted on ChartQA~\cite{masry2022chartqa} test set, CharXiv~\cite{wang2024charxiv} reasoning split, ChartQA-Pro~\cite{masry2025chartqapro}, and TableVQA~\cite{kim2024tablevqabench}.
\end{itemize}
Except for the orientation and resizing augmentations applied to DocVQA and InfoVQA, all other evaluation benchmarks are used in their standard settings.

\paragraph{Implementation Details.}
We initialize our model from Qwen2.5-VL-7B-Instruct~\cite{bai2025qwen25vl} and train with Group Relative Policy Optimization (GRPO)~\cite{shao2024deepseekmath}, sampling $16$ trajectories per input. Training is performed for 200 steps per stage with a learning rate $1{\times}10^{-6}$, batch size 256, ratio clipping $0.2$, and no KL penalty. Each trajectory allows up to 10 tool-use turns. For zoom-in rewards, we utilize IoU threshold to $0.5$ and set $w_{\text{fp}} = 0.1$ and $w_{\text{fn}} = 1$. LLM-Judge in equation (\ref{eq:R_answer}) is obtained using Qwen2.5-VL-72B~\cite{bai2025qwen25vl} and prompted for binary decisions with temperature 0.3. Training is conducted on 4 nodes of $8\times$H200 GPUs with FSDP.

\paragraph{Evaluation Metrics.}
Following~\cite{zheng2025deepeyes,ni2025pointrft,su2025openthinkimg,wu2025vtoolr1}, we report accuracy for all of our evaluation benchmarks except DocVQA-RF, InfoVQA-RF and  InfoVQA-Res. Accuracy is computed using Qwen2.5-VL-72B~\cite{bai2025qwen25vl} as an LLM judge to evaluate answer correctness. For DocVQA-RF, InfoVQA-RF and  InfoVQA-Res we instead report ANLS score following~\cite{mathew2021docvqa,mathew2022infographicvqa}.

\subsection{Experimental Results}

\subsubsection{Main Results}

Table~\ref{tab:main} presents the performance comparison of our proposed ToolsRL against SOTA across three regimes. Our method consistently achieves SOTA performance across the majority of evaluated benchmarks, demonstrating the efficacy and generalizability for tool-use in complex reasoning tasks. On document understanding, ToolsRL achieves a significant lead: $77.3\%$ on DocVQA-RF and $61.4\%$ on InfoVQA-RF. These scores surpass DeepEyes by substantial margins. On spatial understanding, ToolsRL demonstrates strong overall performance on HR-Bench together with clear gains on V-Star and InfoVQA-Res, marking 4.3 and 12.8 point improvements over Mini-o3 on the latter two benchmarks. On Chart/Table understanding, ToolsRL consistently outperforms all other approaches with notable improvements. These outcomes validate tool-supervised RL as an effective paradigm for multi-domain reasoning.
We provide additional qualitative results and analyses (e.g., comparison case study and tool usage comparison) in the supplementary material.

\subsubsection{Ablation Studies}

\paragraph{Reward Design and Curriculum.} To better understand the contributions of each component in ToolsRL, we perform an ablation study covering different reward designs and training strategies, as summarized in Table \ref{tab:ablation_curriculum}. First, introducing the answer reward alone substantially improves performance over the base Qwen2.5-VL-7B model on all benchmarks, e.g., DocVQA-RF rises from $50.2\%$ to $62.6\%$ and TableVQA from $66.2\%$ to $70.6\%$, demonstrating that reward-driven learning helps the model optimize task outcomes even without explicit tool guidance. On top of it, adding a conditional tool reward ($R_{\text{tool\_cond}}$)~\cite{zheng2025deepeyes} alongside answer reward yields mixed improvements. It increases DocVQA-RF accuracy to $71.1\%$ but slightly reduces InfoVQA-RF performance. Similarly, applying tool-supervised rewards alone without curriculum leads to inconsistent gains. These observations motivate our two-stage curriculum pipeline. Examining global and answer-conditioned tool supervision individually reveals complementary effects: global tool supervision mainly benefits chart understanding tasks, while answer-conditioned tool supervision improves spatial reasoning and certain document understanding metrics. Combining these rewards within the ToolsRL framework fully leverages their complementary strengths. With curriculum training, ToolsRL achieves the highest accuracy across nearly all benchmarks, demonstrating that both tool supervision and staged training are essential for effective multi-step visual reasoning.

\paragraph{Tool-supervised Reward Component Design.}

We ablate key design choices in our tool reward formulation (Table \ref{tab:ablation_components}). \textbf{(1) Zoom-in: False Positive Weight.}
Reducing the false-positive weight from $1.0$ to $0.1$ decreases the penalty for incorrect zoom attempts, encouraging exploration of the zoom-in tool. This change increases VisualProbe accuracy ($42.9\%$ $\rightarrow$ $46.3\%$) and average tool calls (2.13 $\rightarrow$ 3.20), showing more effective use of zoom-in actions. \textbf{(2) Rotate and Flip: Augmented Data Only in Stage 1.} When Stage~1 is trained on a mix of augmented (rotated/flipped) and original images, the model often predicts the original image index as correct, ignoring rotated/flipped images. This occurs because original images are substantially easier to answer correctly, creating a shortcut that undermines tool learning. Restricting Stage~1 to augmented data forces the model to actively detect and correct orientation issues, improving DocVQA-RF accuracy (from $67.1\%$ to $79.4\%$) while reducing excessive tool calls (from 6.98 to 4.26). \textbf{(3) Draw: Continuous v.s. Discrete Reward.} We compare a discrete, threshold-based reward, assigning full credit only when predicted primitives fall within 10 pixels of the ground truth, with a continuous, margin-based reward (Sec.~\ref{sec:tool-supervised-rewards}) that provides graded feedback proportional to prediction accuracy. The continuous reward improves optimization stability and facilitates learning of point/line drawing behaviors, resulting in higher ChartQA-Pro accuracy ($37.9\%$ $\rightarrow$ $39.1\%$) and a modest increase in average tool calls (2.43 $\rightarrow$ 2.65).

\begin{table}[t]
    \centering
    \caption{\textbf{Native tool support and usage across methods.} We show which native visual tools each method supports (\checkmark) and their average tool calls per sample during training. Our ToolsRL approach is the only method that supports the full suite of native tools and achieves significantly higher tool usage (3.4 calls) compared to most prior work (${\leq}1$ call).}
    \label{tab:avg-tool-calls}
    \resizebox{0.95\linewidth}{!}{
    \begin{tabular}{l|ccccc|c}
    \toprule
    \textbf{Method} & \textbf{Zoom-in} & \textbf{Rotate} & \textbf{Flip} & \textbf{Line} & \textbf{Point} & \textbf{Avg tool calls} \\
    \midrule
    DeepEyes~\cite{zheng2025deepeyes} & \checkmark & -- & -- & -- & -- & 1.0 \\
    Mini-o3~\cite{lai2025minio3} & \checkmark & -- & -- & -- & -- & 5.5 \\
    ReVPT~\cite{zhou2025revpt} & \checkmark & -- & -- & -- & -- & 0.6 \\
    Pixel Reasoner~\cite{su2025pixelreasoner} & \checkmark & -- & -- & -- & -- & 0.8 \\
    VTool-R1~\cite{wu2025vtoolr1} & -- & -- & -- & \checkmark & -- & 0.3 \\
    OpenThinkIMG~\cite{su2025openthinkimg} & \checkmark & -- & -- & \checkmark & \checkmark & 0.7 \\
\midrule
    \rowcolor{reddy}\textbf{ToolsRL (Ours)} & \checkmark & \checkmark & \checkmark & \checkmark & \checkmark & \textbf{3.4} \\
    \bottomrule
    \end{tabular}
    }
    \vspace{-2mm}
\end{table}
 \begin{table}[t]
    \centering
    \caption{\textbf{Tool-type distribution and composite usage.} We report tool usage distributions grouped by benchmark categories, and the ratio of cases with composite tool use (mixing multiple tools).}
    \label{tab:tool_dist}
    \resizebox{\columnwidth}{!}{
    \begin{tabular}{l|ccccc|c}
        \toprule
        \textbf{Benchmark group} & \textbf{Zoom-in} & \textbf{Rotate} & \textbf{Flip} & \textbf{Line} & \textbf{Point} & \textbf{Comp. ratio} \\
        \midrule
        Document understanding & 31.0\% & 33.4\% & 33.2\% & 1.9\% & 0.5\% & 98.9\% \\
        Spatial reasoning & 89.8\% & 0.7\% & 4.4\% & 0.0\% & 5.1\% & 82.1\% \\
        Chart understanding & 82.8\% & 0.6\% & 1.7\% & 8.7\% & 6.2\% & 95.6\% \\
        \bottomrule
    \end{tabular}
    }
    \vspace{-2mm}
\end{table}
 
\begin{figure*}[t]
    \centering
    \includegraphics[width=\linewidth]{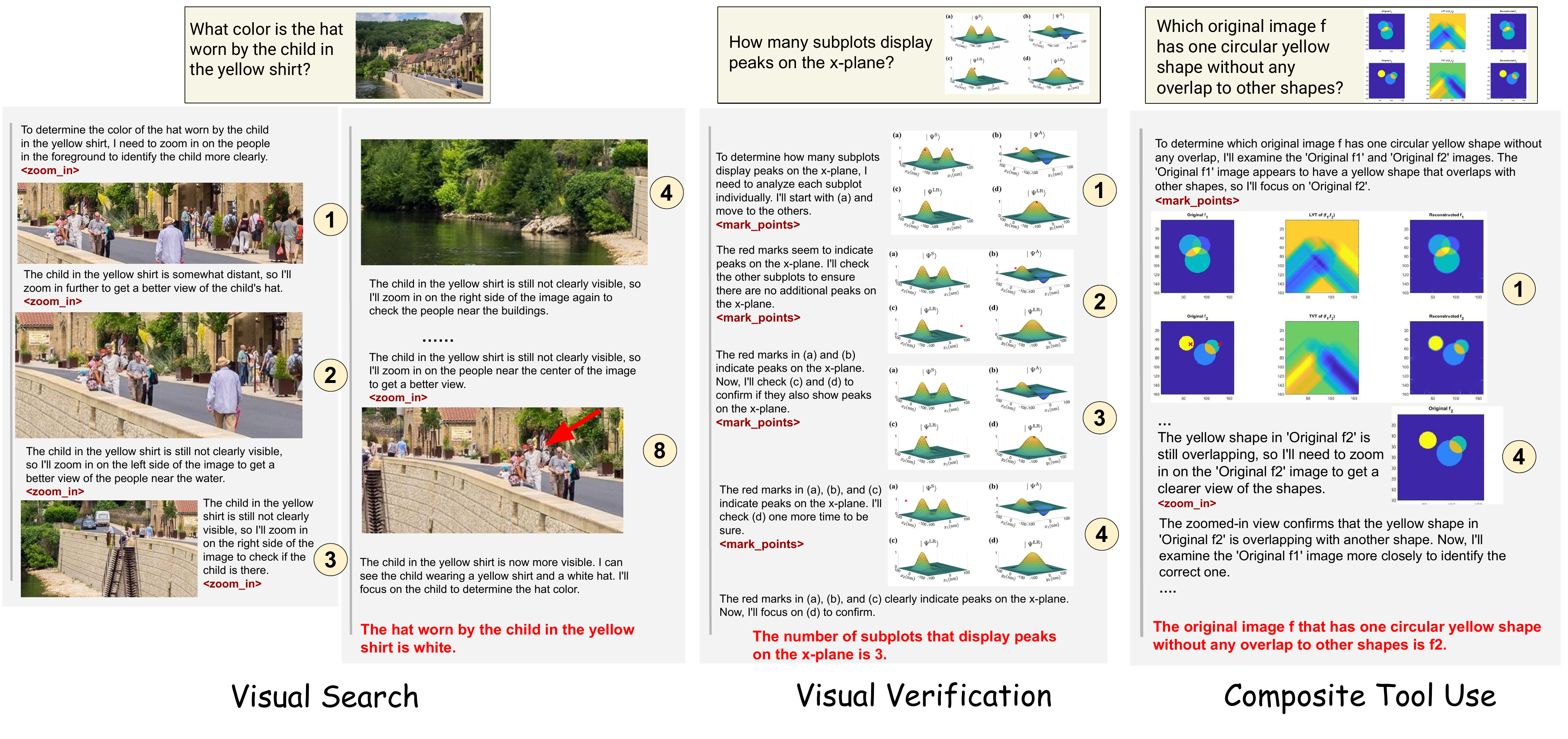}
    \caption{Case studies of ToolsRL. \textbf{Left:} Visual search on high-resolution benchmarks, where the agent iteratively zooms in to localize the queried region before answering. Red arrow in the last image indicates the target region. \textbf{Middle:} Visual verification on charts, where the agent marks key points to check the presence of peaks on the $x$-axis. \textbf{Right:} Composite tool use, where the agent combines zoom-in and point-drawing operations to disambiguate overlapping shapes and identify the correct answer.}
    \label{fig:tool-usage-samples}
\end{figure*}

\begin{figure}[t]
    \centering
    \includegraphics[width=\linewidth]{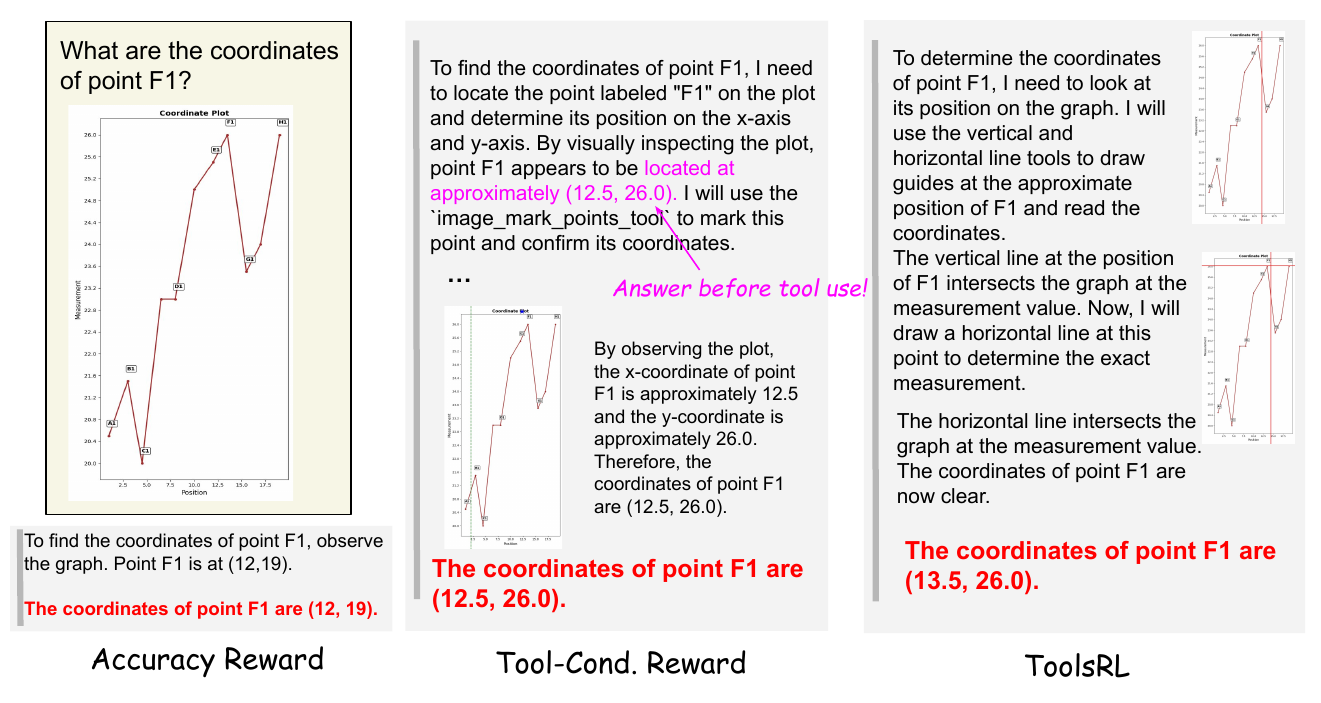}
    \caption{Comparison case study of tool usage across different training settings. The original image and prompt are given in the yellow box, while model answers for different training settings are provided in gray blocks.}
    \label{fig:comparison-case-study}
\end{figure}

\subsection{Results Analysis}

\subsubsection{Native Tool Support and Usage}
Table~\ref{tab:avg-tool-calls} compares native tool support and average tool calls across existing methods and ToolsRL. All prior approaches either support only a single tool or a small set of tools, and most invoke tools rarely (typically $\leq$1 calls per sample, except Mini-o3), relying primarily on text-only reasoning. In contrast, ToolsRL is the only model that supports a wide native toolbox (zoom-in, rotate, flip, draw line, and draw point) and also uses these tools substantially more frequently (averaging 3.4 calls per sample during training).

\paragraph{Tool-type Distribution.}
Table~\ref{tab:tool_dist} further breaks down tool usage by benchmark category. Although each benchmark does exhibit some tool preference, we do not find overly homogeneous tool usage or overly task-specific tool behaviors in general. The high composite ratios across all categories (82--99\%) indicate that ToolsRL learns to combine multiple tools flexibly for complex reasoning.

\subsubsection{Comparison with Different Training Settings}
As illustrated in Figure~\ref{fig:comparison-case-study}, we qualitatively compare how different training strategies shape tool behavior.
Using \textbf{Accuracy Reward Only}, the model often skips tools and directly guesses the answer from the raw image, occasionally issuing a single zoom-in that does not materially change its prediction.
Adding a \textbf{Tool-Conditioned Reward} to the accuracy reward (following~\cite{zheng2025deepeyes}) encourages more frequent tool usage. However, we observe that these tool calls can be noisy or redundant. The agent often outputs an answer before using the tools, indicating reward hacking in the training.
In contrast, our \textbf{Tool-supervision Curriculum} produces reasonable tool trajectories that help the model answer the question correctly.
Additional qualitative comparisons with baselines are provided in the supplementary material.

\subsubsection{Self-Learned Reasoning Patterns}
ToolsRL naturally produces long, tool-driven traces that interleave visual search, measurement, and verification using its full toolbox (zoom-in, rotate, flip, line, and point).
Figure~\ref{fig:tool-usage-samples} visualizes three representative behaviors.
ToolsRL can perform \emph{multi-step visual search}: it progressively zooms into promising regions, examines local evidence, and refines its hypothesis before answering.
On chart domain, it can conduct \emph{visual verification} using pointing to explicitly mark candidate peaks and verify whether they lie on the queried axis.
Finally, it also shows \emph{composite tool use}, chaining zoom-in and draw-point tool calls to resolve ambiguous shapes and reason about occlusions.
We observe trajectories with up to 8 tool calls that still converge to correct answers, indicating that the agent has learned stable, compositional tool-use policies. Importantly, these behaviors emerge purely from our tool-supervision signals instead of complete tool-use trajectories.

 \section{Conclusion}
\label{sec:conclusion}

We introduce \emph{ToolsRL}, a two-stage tool-supervised RL curriculum that decouples tool mastery from answer optimization. In Stage~1, the model learns tool behaviors from ground-truth-derived, per-tool rewards; in Stage~2, it optimizes answer accuracy with GRPO while freely invoking the learned tools. Across document understanding, spatial reasoning, and chart/table understanding, this curriculum yields more stable training, higher accuracy, and stronger visual tool-use patterns than existing methods without requiring expensive, curated tool-use trajectories.
The core ideas—densifying credit assignment via process-level rewards and using a staged curriculum—apply beyond visual tools (e.g., code generation, embodied agents).
 
{
    \small
    \bibliographystyle{ieeenat_fullname}

}

\clearpage
\appendix
{\small
\twocolumn[{
\begin{center}
    \includegraphics[width=0.85\linewidth]{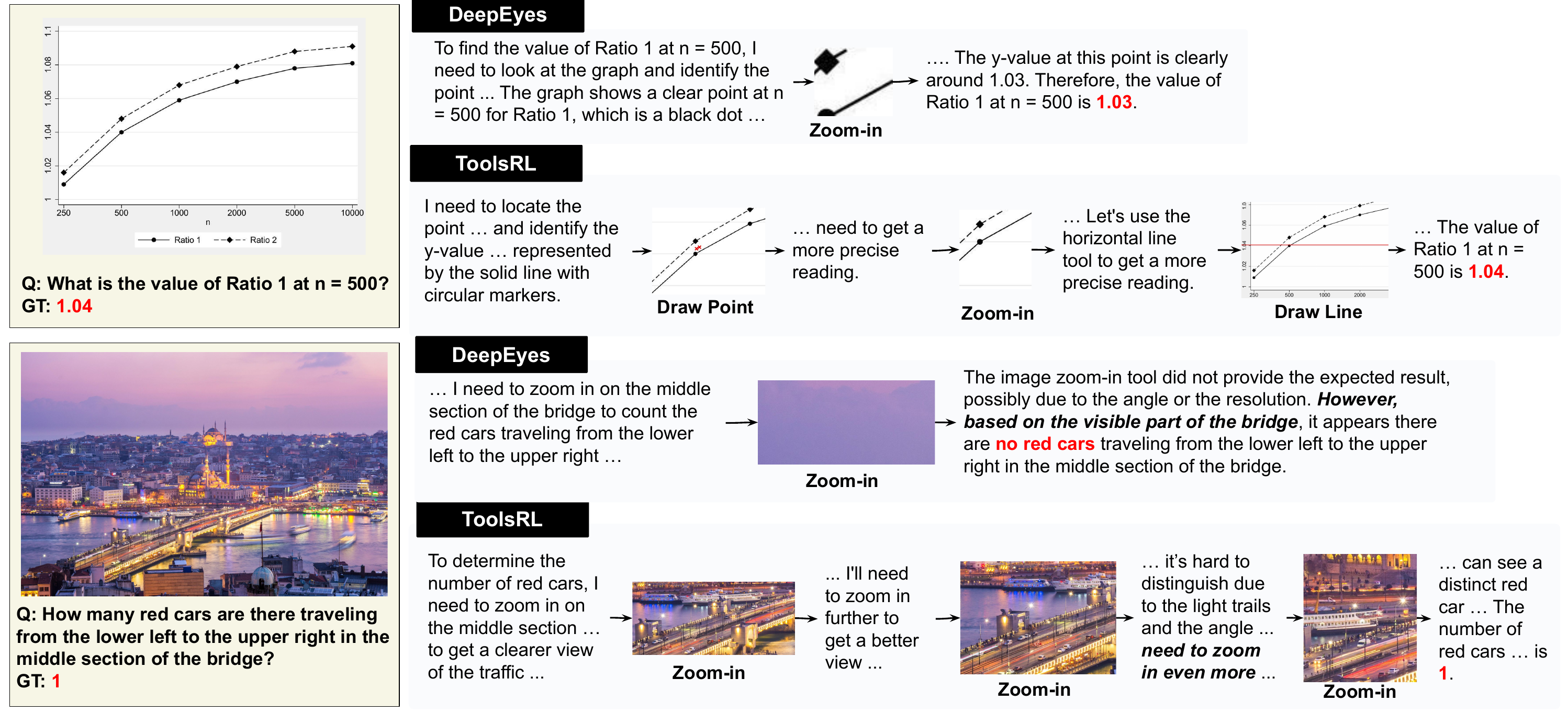}
    \captionof{figure}{Qualitative comparison with DeepEyes.}
    \label{fig:rebuttal_compare_deepeyes}
\end{center}
}]
\section{Overview}
Here we provide additional qualitative comparison (Sec.~\ref{sec:additional_qualitative_comparison}), details on our synthetic dataset generation (Sec.~\ref{sec:dataset_details}), tool-supervised design and ablations (Sec.~\ref{sec:reward_details}), curriculum and ablations (Sec.~\ref{sec:curriculum_ablations}), and the prompts and tool APIs used in both stages of training (Sec.~\ref{sec:prompt_tool_details}).

\section{Qualitative Comparison}
\label{sec:additional_qualitative_comparison}

More qualitative results (ToolsRL vs. DeepEyes) are displayed in Fig.~\ref{fig:rebuttal_compare_deepeyes}, clearly showing ToolsRL is able to use tools more compositely as well as more effectively and precisely.

\begin{figure*}[ht]
    \centering
    \includegraphics[width=\linewidth]{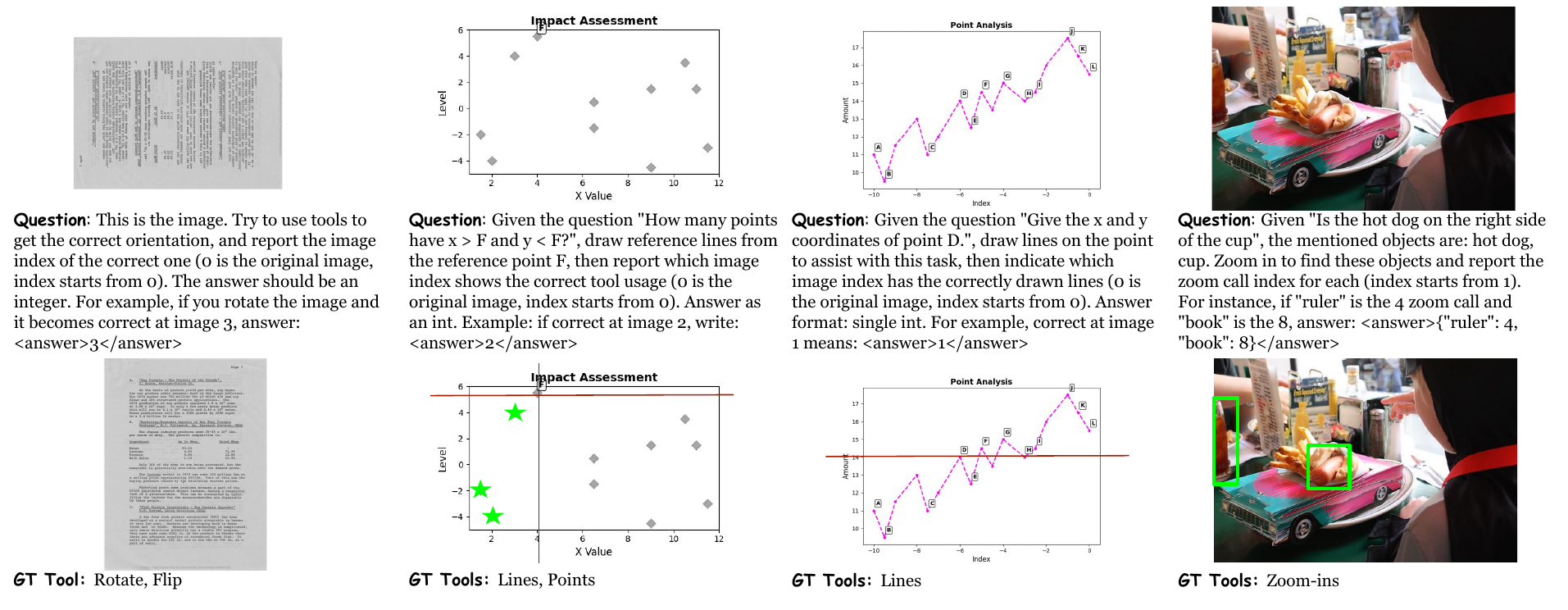}
    \caption{Visual examples of Stage~1 tool-supervised training data. From left to right: (a) document Rotate/Flip orientation correction; (b) chart Read-Value with reference lines; (c) chart Compare-and-Count with marked points and a threshold line; and (d) zoom-in object localization. In each column, the top row shows the question and initial image presented to the model, while the bottom row overlays the ground-truth tool supervision (target orientation, reference lines and points, or zoom-in box) used to compute the rewards.}
    \label{fig:stage1_tool_examples}
\end{figure*}
\section{Synthetic Dataset Generation Details}
\label{sec:dataset_details}

\subsection{Augmented Document Datasets}
We follow the same augmentation pipeline for training and evaluation.

\paragraph{DocVQA-RF and InfoVQA-RF.}
As described in the main paper (Sec.~4), DocVQA and InfoVQA images are rotated by $90^\circ/180^\circ/270^\circ$ or flipped horizontally/vertically, with a transformation applied with probability $0.7$ and sampled uniformly from this set. The \emph{same} rotation/flip distribution is used for both the 3k augmented DocVQA training subset and for constructing the DocVQA-RF and InfoVQA-RF evaluation benchmarks, ensuring that orientation statistics match between training and test.

\paragraph{InfoVQA-Res.}
For InfoVQA-Res, we again follow the construction in Sec.~4: we select InfoVQA images whose maximum edge length exceeds $1024$ pixels and resize them so that the maximum dimension is at most $512$ pixels while preserving aspect ratio.

\subsection{Synthetic Chart Datasets}
We generate two synthetic chart datasets, \textbf{Read-Value} and \textbf{Compare-and-Count}, designed to provide unambiguous ground-truth supervision for drawing tools. These datasets are generated programmatically to ensure precise knowledge of data point locations and values.

\paragraph{Read-Value.}
This dataset consists of synthetically generated scatter and line charts, each paired with a small number of question--answer pairs. Charts are rendered at up to 768\,px on the longer edge, with axis ranges spanning moderate numeric intervals to keep coordinates readable.
\begin{itemize}
    \item \textbf{Task:} Questions ask for the $x$-coordinate, $y$-coordinate, or full $(x, y)$ coordinates of a labeled point (e.g., ``What is the $y$-value of point B?''). Axis-aware variants reuse the chart titles and axis labels to make prompts natural.
    \item \textbf{Chart Design:} Labeled points are placed on scatter or polyline plots with diverse color/marker styles. Each label (letters, numbers, or alphanumeric IDs) is positioned with small offsets so text does not occlude the point, mirroring real chart layouts.
    \item \textbf{Ground Truth and Tool Supervision:} For every labeled point, we store both data coordinates and pixel coordinates on the rendered image. During training, the model is supervised to use \texttt{DrawLine} to draw a horizontal or vertical line from the point to the corresponding axis; this provides precise supervision for where the line should touch the axis and enables accurate reward computation for both the drawn primitive and the numeric answer.
\end{itemize}

\paragraph{Compare-and-Count.}
This dataset contains synthetically generated scatter and line charts, each paired with one comparison question. Images are rendered up to 512\,px on the long edge, and for each question we re-render the plot so that only the single \emph{reference} point is visibly labeled.
\begin{itemize}
    \item \textbf{Task:} Questions ask how many other points satisfy a relational condition with respect to the reference point, such as $x{>}x_{\text{ref}}$, $y{<}y_{\text{ref}}$, both-axes comparisons, or mixed conditions (e.g., ``How many points have $x$ greater than F and $y$ less than F?'').
    \item \textbf{Chart Design:} Under the hood, each chart contains 8--20 rounded points spanning moderate axis ranges, with labels shuffled across letters, numbers, or alphanumeric IDs to avoid positional shortcuts. Although only the reference label is shown in the rendered image, we retain full coordinate metadata for all points.
    \item \textbf{Ground Truth and Tool Supervision:} For every question, we precompute which points qualify under the comparison rule and store their coordinates and labels. During training, the model is encouraged to use \texttt{DrawPoint} to mark qualifying points and \texttt{DrawLine} to indicate threshold boundaries when appropriate (e.g., a vertical line at $x{=}x_{\text{ref}}$). This setup yields dense supervision for both counting behavior and precise spatial localization of the counted set.
\end{itemize}
These synthetic tasks serve as a curriculum capability designed to teach the model precise spatial manipulation and visual working memory usage before it attempts more complex, real-world chart reasoning tasks in Stage 2. Figure~\ref{fig:stage1_tool_examples} shows representative Stage~1 supervision examples across document-rotation, chart reading/counting, and zoom-in tasks.

\section{Tool-Supervised Design and Ablation Details}
\label{sec:reward_details}

\subsection{Zoom-in: ModF1 Reward and Ablation}
Zoom-in in our setting is \emph{not} an object detection task: we only need the zoom window to comfortably cover the region of interest, not to produce a tight bounding box. We therefore adopt a Modified F1 (ModF1) score that down-weights false positives with $w_{\text{fp}}{=}0.1$ while keeping $w_{\text{fn}}{=}1.0$, so missing the target area is penalized much more than including extra background. As illustrated in Fig.~\ref{fig:modf1_ablation}, this reward gives full credit to generous crops that contain the GT box, whereas a symmetric F1 would assign a low score and discourage such safe zoom behavior.

\begin{figure}[th]
    \centering
    \includegraphics[width=0.8\linewidth]{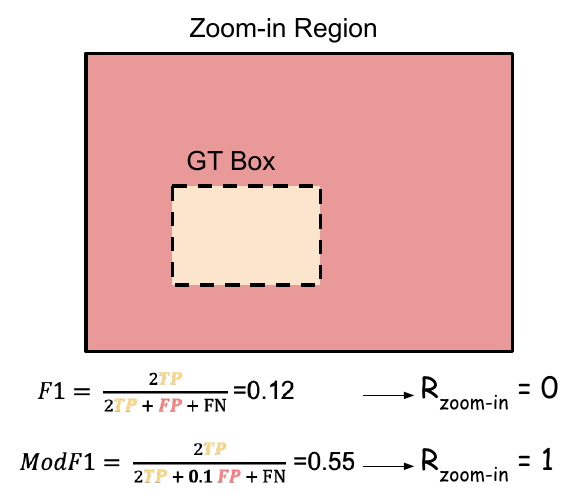}
    \caption{Illustration of standard F1 vs.\ ModF1 for zoom-in. In this example, the predicted zoom region fully contains the ground-truth (GT) box but is much larger. Standard F1 yields a low score (and thus $R_{\text{zoom-in}}{=}0$) because the FP area dominates. ModF1, with a smaller FP weight ($w_{\text{fp}}{=}0.1$), assigns a much higher score (and $R_{\text{zoom-in}}{=}1$), reflecting that generous crops that include target area should still receive full credit.}
    \label{fig:modf1_ablation}
\end{figure}

\subsection{Draw: Discrete vs.\ Continuous Rewards}
For \texttt{DrawLine} and \texttt{DrawPoint} tasks, we compare a simple discrete reward against the continuous margin-based reward used in our main draw metric, both defined in terms of the same distance $d(u,g)$ and tolerance $T_g$ between a predicted primitive $u$ and a ground-truth primitive $g$ in Sec.~3.2.

The discrete reward only gives credit when the prediction lands inside the tolerance window:
\begin{equation}
    R_{\text{discrete}}(u, g) = \mathbb{1}\bigl(d(u,g) < 10\bigr),
\end{equation}
while the continuous version scales linearly with distance:
\begin{equation}
    s(u, g) = \max\!\Bigl(0,\, 1 - \tfrac{d(u,g)}{T_g}\Bigr).
\end{equation}
The discrete signal is too sparse: at the start of training the model almost never discovers useful draw behavior, and the average usage of the point-marking tool stays extremely low (0.23 calls per sample). After switching to the continuous reward (which gives partial credit to near misses), the model begins to actively explore drawing, and the average mark-point usage rises to 0.643. This empirical gap shows that more informative, continuous rewards are essential for learning reliable draw-tool usage.

\subsection{Rotate and Flip: Training with Mixed Orientations}
As observed in our ablation study in Tab.~3, we find it crucial to use \textit{only} the augmented (rotated/flipped) samples in Stage 1 training. If the training mix includes many canonical (un-augmented) documents, the model learns a shortcut: it predicts the answer directly assuming the image is upright, as this strategy works for the canonical subset (which is often easier to answer). By restricting Stage 1 to only rotated/flipped examples, we force the model to use the \texttt{Rotate} or \texttt{Flip} tools to recover the readable orientation before answering. This enforced active perception prevents reward hacking and ensures the tool-use policy is robustly learned.
Figure~\ref{fig:rotflip_ablation} illustrates this phenomenon: when canonical documents are included in the training mix, the model shortcuts by predicting answer index 0 directly; training exclusively on rotated/flipped samples forces active tool use and leads to robust behavior.

\begin{figure}[th]
    \centering
    \includegraphics[width=\linewidth]{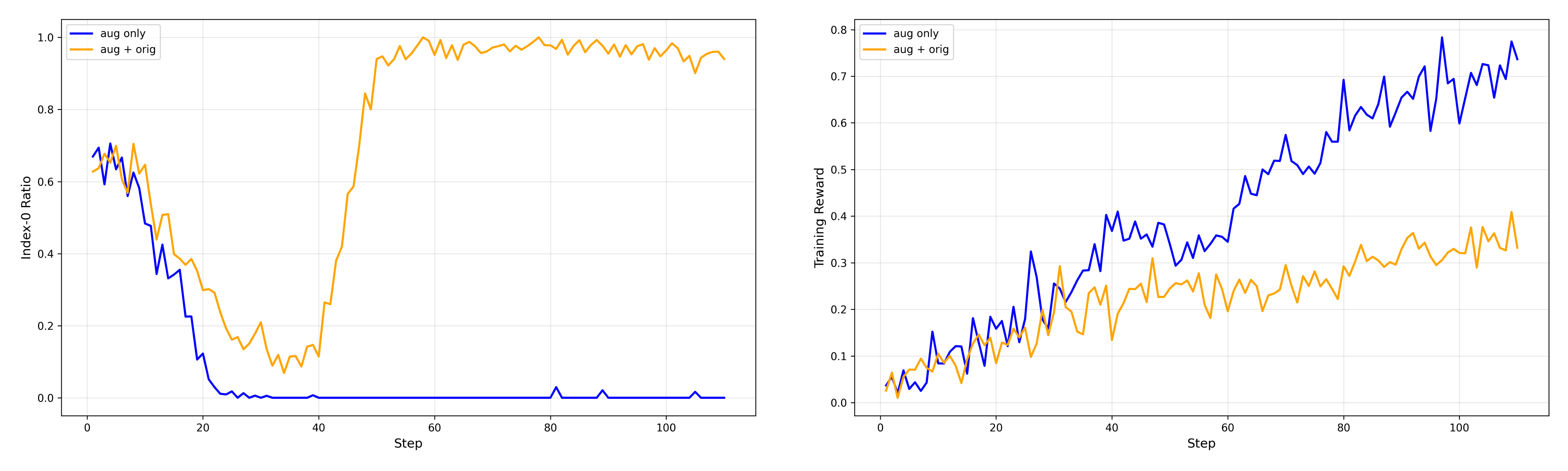}
    \caption{Analysis of the Stage-1 rotation/flip training design. Mixing original and augmented documents (``aug + orig'') encourages a shortcut where the model always predicts index 0 in answer, while training on augmented-only samples (``aug only'') removes this shortcut and yields higher reward.}
    \label{fig:rotflip_ablation}
\end{figure}

\section{Curriculum and Ablation Details}
\label{sec:curriculum_ablations}

\subsection{Overview of Curricula and Baselines}
Table~2 in the main paper compares a family of training strategies: (i) \emph{Accuracy Reward Only}, which optimizes answer correctness without any explicit tool signal; (ii) \emph{Tool-Conditioned Reward}, which adds a scalar bonus when tools are used on correctly answered trajectories; and (iii) our \emph{Tool-supervision Curricula}, where Stage~1 is trained with global and/or answer-conditioned tool rewards and Stage~2 uses only the answer-accuracy reward.

\paragraph{Tool-Conditioned Reward (DeepEyes baseline).}
Following the DeepEyes setup, we define a binary tool-conditioned bonus on each trajectory $\tau$ using the answer reward $R_{\text{answer}}(\tau)$ and the total number of tool calls $\text{tool\_count}(\tau)$:
\begin{equation}
\resizebox{0.85\linewidth}{!}{$
    R_{\text{tool\_cond}}(\tau)
    = \mathbb{1}\!\bigl[R_{\text{answer}}(\tau) > 0.5\bigr]\cdot
      \mathbb{1}\!\bigl[\text{tool\_count}(\tau) > 0\bigr]
$}
\end{equation}
where $\text{tool\_count}(\tau)$ counts all native visual-tool API calls.
The DeepEyes baseline then optimizes
\begin{equation}
\resizebox{0.85\linewidth}{!}{$
    R_{\text{DeepEyes}}(\tau) = R_{\text{answer}}(\tau) + R_{\text{format}}(\tau) + R_{\text{tool\_cond}}(\tau)
$}
\end{equation}
so tools are rewarded only when the final answer is already correct, without any guidance on which tools to invoke or how to structure tool trajectories.

\subsection{Training Dynamics vs Tool-Conditioned Reward}
Figure~\ref{fig:training_curves_and_tool_usage} (left) compares the average number of tool calls per step during Stage~2 training for Accuracy Reward Only, Tool-Conditioned Reward, and our ToolsRL curriculum.
All three settings are trained on the same data with identical prompts and the same answer-accuracy objective; the only difference is whether having Stage~1 for tool supervision.
Accuracy Reward Only almost never invokes tools (staying near one call per sample), and Tool-Conditioned Reward increases tool usage only modestly.
In contrast, the model trained with our Tool-supervision Curriculum consistently uses tools much more frequently (around 3--5 calls per sample) even though Stage~2 has \emph{no explicit tool bonus}, indicating that Stage~1 tool rewards induce a persistent, tool-centric reasoning policy rather than transient reward hacking.

\subsection{Tool Call Rates of Ablated Curricula}
Figure~\ref{fig:training_curves_and_tool_usage} (right) analyzes Stage~1 ablations of our curriculum.
Using only answer-conditioned tool supervision $R^{\text{answer}}_{\text{tool}}$ yields relatively low tool counts: the agent learns to be conservative and invokes tools only when they directly affect the final answer, which limits exploration.
In contrast, using only global tool supervision $R^{\text{global}}_{\text{tool}}$ leads to very high call rates and many redundant actions, since any tool call on the trajectory is rewarded regardless of its usefulness, encouraging inefficient behavior.
Our full ToolsRL curriculum, which combines global and answer-conditioned rewards, lies between these extremes: it promotes rich exploration early in training while gradually shaping the policy toward efficient, task-relevant tool usage.

\begin{figure}[t]
    \centering
    \includegraphics[width=\linewidth]{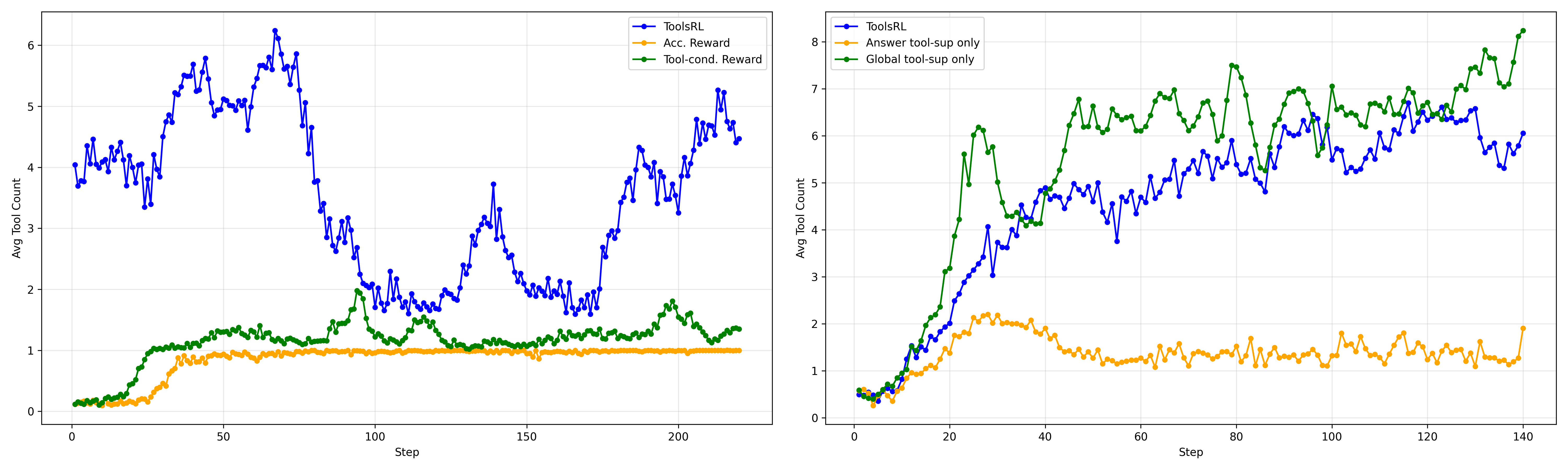}
    \caption{Average tool usage during training across ablation experiments. \textbf{Left:} comparison between Accuracy Reward Only, Tool-Conditioned Reward, and our ToolsRL during Stage~2, all trained on the same data, prompts, and answer-accuracy objective. \textbf{Right:} Stage~1 average tool calls for our full curriculum versus answer-only and global-only tool supervision.}
    \label{fig:training_curves_and_tool_usage}
\end{figure}

\section{Prompt and Tool Argument Details}
\label{sec:prompt_tool_details}

\subsection{Stage~1 Tool-supervised Prompts}
Stage~1 uses task-specific system prompts that expose only the tools needed for each supervision regime, paired with a shared lightweight user instruction.
The literal system prompts are shown in Tables~\ref{tab:zoom_prompt}--\ref{tab:read_value_prompt}; for tool-argument details, refer to the Tool API in Sec.~\ref{sec:tool_api}.
\begin{itemize}
    \item \textbf{Read-Value and Compare-and-Count.} The system prompt advertises the point and line tools---\texttt{image\_mark\_points\_tool}, \texttt{image\_draw\_horizontal\_line\_tool}, and \texttt{image\_draw\_vertical\_line\_tool}---and emphasizes that all coordinates must be given in pixel space on the rendered chart (image columns/rows), not axis values.
    \item \textbf{Rotate/Flip tasks.} The system prompt exposes \texttt{image\_rotate\_tool} and \texttt{image\_flip\_tool} and explains that angles are in degrees (positive = clockwise) and flips are either horizontal or vertical.
    \item \textbf{Zoom-in tasks.} The system prompt exposes only \texttt{image\_zoom\_in\_tool} with a bounding-box interface for cropping.
\end{itemize}

All Stage~1 datasets share the same concise user prompt:
\begin{tcolorbox}[title=Shared User Prompt (Stage~1 and Stage~2), colback=white, colframe=nicegray, fonttitle=\bfseries]
\footnotesize
\begin{lstlisting}[breakindent=0pt]
{Question}
Think first, call tools if needed, then answer.
Format as: \texttt{<think>...</think> <tool_call>...</tool_call>} (if tools needed) or \texttt{<think>...</think> <answer>...</answer>} (if you know the answer).
\end{lstlisting}
\end{tcolorbox}
This enforces a consistent trace structure across zoom, rotate/flip, and drawing tasks without leaking reward details.

\begin{table*}[p]
    \centering
    \caption{System Prompt for Zoom-in Task (Stage 1; abbreviated tool-argument text, see Tool API in Sec.~\ref{sec:prompt_tool_details} for full definitions).}
    \label{tab:zoom_prompt}
    \begin{tabular}{p{0.98\linewidth}}
    \toprule
    \begin{lstlisting}[basicstyle=\ttfamily\scriptsize, breaklines=true, breakindent=0pt, aboveskip=2pt, belowskip=2pt]
You are a helpful assistant.

# Tools
You may call one or more functions to assist with the user query.
You are provided with function signatures within <tools></tools> XML tags:
<tools>
{"type":"function","function":{"name":"image_zoom_in_tool","description":"Zoom in on a specific region...","parameters":{"type":"object","properties":{"bbox_2d":{"type":"array","items":{"type":"number"},"minItems":4,"maxItems":4,"description":"The bounding box of the region to zoom in..."},"label":{"type":"string","description":"..."},"target_image":{"type":"integer","description":"...","default":-1}},"required":["bbox_2d"]}}
</tools>

# How to call a tool
Return a json object with function name and arguments within <tool_call></tool_call> XML tags:
<tool_call>
{"name": <function-name>, "arguments": <args-json-object>}
</tool_call>

**Example**:
<tool_call>
{"name": "image_zoom_in_tool", "arguments": {"bbox_2d": [50, 100, 200, 300], "label": "the object in the center", "target_image": 1}}
</tool_call>
    \end{lstlisting} \\
    \bottomrule
    \end{tabular}
\end{table*}

\begin{table*}[p]
    \centering
    \caption{System Prompt for Rotate/Flip Tasks (Stage 1; abbreviated tool-argument text, see Tool API in Sec.~\ref{sec:prompt_tool_details} for full definitions).}
    \label{tab:rot_flip_prompt}
    \begin{tabular}{p{0.98\linewidth}}
    \toprule
    \begin{lstlisting}[basicstyle=\ttfamily\scriptsize, breaklines=true, breakindent=0pt, aboveskip=2pt, belowskip=2pt]
You are a helpful assistant.

# Tools
You may call one or more functions to assist with the user query.
You are provided with function signatures within <tools></tools> XML tags:
<tools>
{"type":"function","function":{"name":"image_rotate_tool","description":"Rotate an image by a specified angle...","parameters":{"type":"object","properties":{"angle":{"type":"integer","description":"Rotation angle in degrees..."},"label":{"type":"string","description":"..."},"target_image":{"type":"integer","description":"...","default":-1}},"required":["angle"]}}
{"type":"function","function":{"name":"image_flip_tool","description":"Flip an image horizontally or vertically...","parameters":{"type":"object","properties":{"direction":{"type":"string","enum":["horizontal","vertical"],"description":"Direction to flip..."},"label":{"type":"string","description":"..."},"target_image":{"type":"integer","description":"...","default":-1}},"required":["direction"]}}
</tools>

# How to call a tool
Return a json object with function name and arguments within <tool_call></tool_call> XML tags:
<tool_call>
{"name": <function-name>, "arguments": <args-json-object>}
</tool_call>

**Examples**:
<tool_call>
{"name": "image_rotate_tool", "arguments": {"angle": 270, "label": "rotate image for better view", "target_image": 2}}
</tool_call>
<tool_call>
{"name": "image_flip_tool", "arguments": {"direction": "horizontal", "label": "flip image to correct orientation", "target_image": 1}}
</tool_call>
    \end{lstlisting} \\
    \bottomrule
    \end{tabular}
\end{table*}

\begin{table*}[p]
    \centering
    \caption{System Prompt for Read-Value and Compare-and-Count Tasks (Stage 1; abbreviated tool-argument text, see Tool API in Sec.~\ref{sec:prompt_tool_details} for full definitions).}
    \label{tab:read_value_prompt}
    \begin{tabular}{p{0.98\linewidth}}
    \toprule
    \begin{lstlisting}[basicstyle=\ttfamily\scriptsize, breaklines=true, breakindent=0pt, aboveskip=2pt, belowskip=2pt]
You are a helpful assistant.

# Tools
You may call one or more functions to assist with the user query.
You are provided with function signatures within <tools></tools> XML tags:
<tools>
{"type":"function","function":{"name":"image_mark_points_tool","description":"Mark specific points on an image...","parameters":{"type":"object","properties":{"point_2d":{"type":"array","description":"Use pixel coordinates..."},"color":{"type":"string","description":"..."},"size":{"type":"integer","description":"..."},"shape":{"type":"string","enum":["X","star"],"description":"..."},"label":{"type":"string","description":"..."},"target_image":{"type":"integer","description":"...","default":-1}},"required":["point_2d"]}}
{"type":"function","function":{"name":"image_draw_horizontal_line_tool","description":"Draw a horizontal line...","parameters":{"type":"object","properties":{"height_location":{"type":"integer","description":"Image row index..."},"color":{"type":"string","description":"..."},"thickness":{"type":"integer","description":"..."},"style":{"type":"string","enum":["solid","dashed"],"description":"..."},"label":{"type":"string","description":"..."},"target_image":{"type":"integer","description":"...","default":-1}},"required":["height_location"]}}
{"type":"function","function":{"name":"image_draw_vertical_line_tool","description":"Draw a vertical line...","parameters":{"type":"object","properties":{"width_location":{"type":"integer","description":"Image column index..."},"color":{"type":"string","description":"..."},"thickness":{"type":"integer","description":"..."},"style":{"type":"string","enum":["solid","dashed"],"description":"..."},"label":{"type":"string","description":"..."},"target_image":{"type":"integer","description":"...","default":-1}},"required":["width_location"]}}
</tools>

# How to call a tool
Return a json object with function name and arguments within <tool_call></tool_call> XML tags:
<tool_call>
{"name": <function-name>, "arguments": <args-json-object>}
</tool_call>

**Examples**:
<tool_call>
{"name": "image_mark_points_tool", "arguments": {"point_2d": [[253, 375], [190, 220]], "color": "yellow", "size": 8, "shape": "star", "label": ["Point A", "Point B"], "target_image": -1}}
</tool_call>
<tool_call>
{"name": "image_draw_horizontal_line_tool", "arguments": {"height_location": 157, "color": "purple", "thickness": 3, "style": "dashed", "label": "horizontal guide at this height location", "target_image": 0}}
</tool_call>
<tool_call>
{"name": "image_draw_vertical_line_tool", "arguments": {"width_location": 209, "color": "blue", "thickness": 4, "style": "solid", "label": "vertical guide at this width location", "target_image": 0}}
</tool_call>

Always remember: location values such as the `[width_location, height_location]` pairs in `point_2d` and the separate `width_location` / `height_location` arguments for line tools are pixel measurements from the rendered image (width = columns, height = rows), not the chart-axis numbers. Match the screenshot pixels: if a chart point is labelled (15,10) but appears at pixel column 157 and row 200, call the tools with `point_2d`: [157, 200], `width_location`: 157, and `height_location`: 200.
    \end{lstlisting} \\
    \bottomrule
    \end{tabular}
\end{table*}

\begin{table*}[p]
    \centering
    \caption{System Prompt for Stage 2 QA (abbreviated tool-argument text, see Tool API in Sec.~\ref{sec:prompt_tool_details} for full definitions).}
    \label{tab:stage2_prompt}
    \begin{tabular}{p{0.98\linewidth}}
    \toprule
    \begin{lstlisting}[basicstyle=\ttfamily\scriptsize, breaklines=true, breakindent=0pt, aboveskip=2pt, belowskip=2pt]
You are a helpful assistant.

# Tools
You may call one or more functions to assist with the user query.
You are provided with function signatures within <tools></tools> XML tags:
<tools>
{"type":"function","function":{"name":"image_mark_points_tool","description":"Mark specific points on an image...","parameters":{"type":"object","properties":{"point_2d":{"type":"array","description":"Use pixel coordinates..."},"color":{"type":"string","description":"..."},"size":{"type":"integer","description":"..."},"shape":{"type":"string","enum":["circle","X","star"],"description":"..."},"label":{"type":"string","description":"..."},"target_image":{"type":"integer","description":"...","default":-1}},"required":["point_2d"]}}
{"type":"function","function":{"name":"image_zoom_in_tool","description":"Zoom in on a specific region...","parameters":{"type":"object","properties":{"bbox_2d":{"type":"array","items":{"type":"number"},"minItems":4,"maxItems":4,"description":"Bounding box as [x1, y1, x2, y2]..."},"label":{"type":"string","description":"..."},"target_image":{"type":"integer","description":"...","default":-1}},"required":["bbox_2d"]}}
{"type":"function","function":{"name":"image_draw_horizontal_line_tool","description":"Draw a horizontal line...","parameters":{"type":"object","properties":{"height_location":{"type":"integer","description":"Image row index..."},"color":{"type":"string","description":"..."},"thickness":{"type":"integer","description":"..."},"style":{"type":"string","enum":["solid","dashed"],"description":"..."},"label":{"type":"string","description":"..."},"target_image":{"type":"integer","description":"...","default":-1}},"required":["height_location"]}}
{"type":"function","function":{"name":"image_draw_vertical_line_tool","description":"Draw a vertical line...","parameters":{"type":"object","properties":{"width_location":{"type":"integer","description":"Image column index..."},"color":{"type":"string","description":"..."},"thickness":{"type":"integer","description":"..."},"style":{"type":"string","enum":["solid","dashed"],"description":"..."},"label":{"type":"string","description":"..."},"target_image":{"type":"integer","description":"...","default":-1}},"required":["width_location"]}}
{"type":"function","function":{"name":"image_rotate_tool","description":"Rotate an image by a specified angle...","parameters":{"type":"object","properties":{"angle":{"type":"integer","description":"Rotation angle in degrees..."},"label":{"type":"string","description":"..."},"target_image":{"type":"integer","description":"...","default":-1}},"required":["angle"]}}
{"type":"function","function":{"name":"image_flip_tool","description":"Flip an image horizontally or vertically...","parameters":{"type":"object","properties":{"direction":{"type":"string","enum":["horizontal","vertical"],"description":"Flip direction."},"label":{"type":"string","description":"..."},"target_image":{"type":"integer","description":"...","default":-1}},"required":["direction"]}}
</tools>

# How to call a tool
Return a json object with function name and arguments within <tool_call></tool_call> XML tags:
<tool_call>
{"name": <function-name>, "arguments": <args-json-object>}
</tool_call>

**Examples**:
<tool_call>
{"name": "image_mark_points_tool", "arguments": {"point_2d": [[188, 114], [324, 200]], "color": "green", "size": 6, "shape": "X", "label": ["Point A", "Point B"], "target_image": 2}}
</tool_call>
<tool_call>
{"name": "image_zoom_in_tool", "arguments": {"bbox_2d": [120, 90, 360, 320], "label": "zoom to the bar labels", "target_image": 0}}
</tool_call>
<tool_call>
{"name": "image_draw_horizontal_line_tool", "arguments": {"height_location": 205, "color": "yellow", "thickness": 3, "style": "dashed", "label": "horizontal guide for reading value", "target_image": 0}}
</tool_call>
<tool_call>
{"name": "image_draw_vertical_line_tool", "arguments": {"width_location": 178, "color": "purple", "thickness": 3, "style": "solid", "label": "vertical guide for comparing x-values", "target_image": -1}}
</tool_call>
<tool_call>
{"name": "image_rotate_tool", "arguments": {"angle": -45, "label": "rotate image to upright orientation", "target_image": 1}}
</tool_call>
<tool_call>
{"name": "image_flip_tool", "arguments": {"direction": "horizontal", "label": "flip image to correct orientation", "target_image": 0}}
</tool_call>

Always remember: all coordinates (`bbox_2d`, `height_location`, `width_location`, `point_2d`) must use pixel locations from the rendered image (width = columns, height = rows), not chart-axis values.
    \end{lstlisting} \\
    \bottomrule
    \end{tabular}
\end{table*}

\subsection{Stage~2 QA Prompts}
Stage~2 switches to a unified QA setting where a single system prompt exposes the full toolbox.
The full Stage~2 system prompt is listed in Table~\ref{tab:stage2_prompt}; for tool-argument details, refer to the Tool API in Sec.~\ref{sec:tool_api}.
The system prompt advertises \texttt{image\_zoom\_in\_tool}, \texttt{image\_rotate\_tool}, \texttt{image\_flip\_tool}, \texttt{image\_draw\_horizontal\_line\_tool}, \texttt{image\_draw\_vertical\_line\_tool}, and \texttt{image\_mark\_points\_tool} together, using the same JSON-style \texttt{<tool\_call>} convention as in Stage~1.
The user prompt is kept identical to Stage~1, so the main change between stages is the reward objective (answer accuracy) and toolbox breadth, not the prompt format.

\subsection{Tool API}
\label{sec:tool_api}
For completeness, we briefly summarize the tool interfaces (including arguments, constraints, and defaults) used during both stages.
\begin{itemize}
    \item \textbf{\texttt{image\_zoom\_in\_tool}}: Takes a 4-number array \texttt{bbox\_2d = [x1, y1, x2, y2]} in pixel coordinates (image width, height).\footnote{Throughout, ``pixel coordinates'' mean column/row indices on the rendered image, not axis values.} Optionally accepts a textual \texttt{label}. The optional \texttt{target\_image} index selects which buffered image to crop: negative values (\texttt{-1}, \texttt{-2}, \dots) denote the last $N$ processed images, positive values index the image lineage (\texttt{0} = original, \texttt{1} = after first tool call, etc.), and the default is \texttt{-1} (last processed image).
    \item \textbf{\texttt{image\_rotate\_tool}}: Requires an integer \texttt{angle} in degrees; positive values rotate clockwise and negative values counterclockwise (e.g., \texttt{90}, \texttt{180}, \texttt{270}). Optionally takes a descriptive \texttt{label} and a \texttt{target\_image} index with the same convention as above (default \texttt{-1}).
    \item \textbf{\texttt{image\_flip\_tool}}: Requires a \texttt{direction} string in \{\texttt{horizontal}, \texttt{vertical}\}, indicating left--right vs.\ top--bottom flips. Optionally accepts a \texttt{label} and \texttt{target\_image} (again defaulting to \texttt{-1} for the last processed image).
    \item \textbf{\texttt{image\_draw\_horizontal\_line\_tool}}: Requires an integer \texttt{height\_location} giving the image row (pixel) where the horizontal line should be drawn. Optional arguments include \texttt{color} (e.g., \texttt{"red"}, \texttt{"blue"}, \texttt{"green"}), \texttt{thickness} (pixels), \texttt{style} in \{\texttt{solid}, \texttt{dashed}\}, a textual \texttt{label}, and \texttt{target\_image} (default \texttt{-1}).
    \item \textbf{\texttt{image\_draw\_vertical\_line\_tool}}: Requires an integer \texttt{width\_location} giving the image column (pixel) where the vertical line should be drawn, with the same optional \texttt{color}, \texttt{thickness}, \texttt{style}, \texttt{label}, and \texttt{target\_image} (default \texttt{-1}) as the horizontal line tool.
    \item \textbf{\texttt{image\_mark\_points\_tool}}: Takes \texttt{point\_2d} as either a single \texttt{[width\_location, height\_location]} pair or an array of such pairs (all in pixel coordinates). Optional arguments include \texttt{color}, \texttt{size} (marker radius in pixels), \texttt{shape} (e.g., \texttt{circle}, \texttt{X}, \texttt{star}), \texttt{label} (single string or list of strings for multiple points), and \texttt{target\_image} (default \texttt{-1}).
\end{itemize}
Across all tools, the prompts consistently remind the model that every coordinate---\texttt{bbox\_2d}, \texttt{height\_location}, \texttt{width\_location}, and \texttt{point\_2d}---must be expressed in the pixel space of the current rendered image, rather than in chart-axis units.
 }

\end{document}